\documentclass{article}

    \PassOptionsToPackage{numbers, compress}{natbib}

\usepackage[dblblindworkshop, final]{neurips_2026}

\usepackage[utf8]{inputenc} 
\usepackage[T1]{fontenc}    
\usepackage{hyperref}       
\usepackage{url}            
\usepackage{booktabs}       
\usepackage{amsfonts}       
\usepackage{nicefrac}       
\usepackage{microtype}      
\usepackage{xcolor}         

\usepackage{authblk}
\usepackage{amsmath}
\usepackage{placeins}
\usepackage{graphicx}
\usepackage{longtable}      

\title{Zero-Compute Cross-Lingual Transferability Estimation Using Typological Feature Proxies}
\workshoptitle{Linguistic Principles for Foundation Models (LP4FM)}

\author[1]{\textbf{Dalton~Raphael~Harmsen\textsuperscript{*}}%
}%
\author[2]{\textbf{Swier~Garst}}
\author[2]{\textbf{Thomas~van~Osch}}
\author[2]{\textbf{Zarè~Palanciyan}}
\author[1]{\textbf{Joaquin~Vanschoren}}

\affil[1]{%
    AMOR/e Lab\\
    Eindhoven University of Technology\\
    Eindhoven, The Netherlands
}
\affil[2]{%
    SURF\\
    Amsterdam, The Netherlands
}

\begin{document}

\begingroup
\renewcommand{\thefootnote}{*}
\endgroup

\maketitle

\begin{abstract}
    Cross-lingual transfer describes how knowledge in a source language benefits a target language.
Measuring it quantitatively requires broad multilingual pre-training, as prior work has done with cross-lingual transfer matrices.
We ask whether transfer is predictable from freely available typological features, and whether the prominence of high-resource source languages reflects typology or data quality and quantity.
We show that typological databases contain cheap and dense signals about cross-lingual transfer.
Our typology-only random forest on a 24-language prior-work transfer matrix scores leave-one-language-out $\rho{=}0.705$ and $R^2{=}0.49$, against $\rho{=}0.62$ for a tuned non-typological control, an increment that a paired bootstrap cannot separate at this scale.
When both languages of a held-out pair are unseen, performance degrades to $\rho{\approx}0.41$, so the model interpolates a ground-truth matrix more reliably than it predicts new languages.
The signal survives leave-one-script-out and leave-one-family-out protocols, so script and family confounding do not explain the effect.
By decomposing the transfer into a typology term and a resource-and-script bias term, we find the best-source ranking sensitive to this bias.
Typological features never observe this bias, which makes them a zero-compute screening tool that replaces hundreds of training runs with a model fit.
Our code is available \href{https://github.com/dharmsen/typo-x-ling-transfer}{here}.

\end{abstract}

\section{Introduction}\label{sec:intro}
Cross-lingual transfer describes how knowledge in a source language benefits a target language. \citeauthor{ICLR2026_35c26a81} introduced ATLAS, which measures this through multilingual pre-training, finding that a few high-resource languages dominate in transfer due to their abundance and quality of training data.
However, methods such as ATLAS are prohibitively expensive, requiring hundreds of pre-training runs~\cite{cao2026shapleylaw,he-etal-2025-scaling}.
On the other hand, Grambank~\cite{skirgardGrambankRevealsImportance2023,grambank_dataset_zenodo_v1} and WALS~\cite{wals} are readily-available typological databases holding hundreds of structural properties per language.
Linguistic similarity has previously been shown to correlate with transfer~\cite{make8030065,pmlr-v203-muller23a,rice-etal-2025-untangling}.
We thus hypothesize that these databases contain valuable transfer signal, which can rank candidate source languages before any compute is spent.
We pose two questions;
\textbf{RQ1}: can typological features reconstruct the transfer scores of the ATLAS matrix?
\textbf{RQ2}: is the prominence of high-resource languages such as English driven by typological features, or by data quality and quantity?
\paragraph{Contributions}
First, we show that a typology-only random forest predicts transfer on a 24-language ATLAS subset, verifying the ability of typology-only predictions to reconstruct the cross-lingual transfer measured by ATLAS.
The contribution is the empirical finding that typological databases carry transfer signal at zero compute.
Our inexpensive screen complements forward-pass estimators such as LEEP~\cite{pmlr-v119-nguyen20b} and LogME~\cite{pmlr-v139-you21b} and replaces hundreds of LLM training runs.
Second, we decompose the ATLAS scores into a typology term and a resource-and-script bias term, and show that best-source ranking is sensitive to this bias.

\section{Method and Experiments}\label{sec:method}
\paragraph{Data}
We use typological features from both the Grambank~\cite{skirgardGrambankRevealsImportance2023,grambank_dataset_zenodo_v1} and WALS~\cite{wals} databases.
In total, 387 Glottocode-keyed~\cite{doi:10.3233/SW-212843} features were used, of which 195 originated from Grambank, and 192 from WALS.
The exact list of features is given in Appendix~\ref{app:config}, Table~\ref{tab:features}.
We use these features to predict the bilingual transfer scores (BTS) and fine-tuning adaptation score (FAS) introduced by ATLAS~\cite{ICLR2026_35c26a81}. Keeping languages with at least 70\% Grambank feature coverage leaves 24 languages and 552 language pairs; a set we refer to as ATLAS-24.
%
\paragraph{Model training}
Our main model is a 200-tree random forest.
We compare four tiers: two random forest models, one using all features and one using the features determined by a feature selection and hyperparameter sweep, and two ridge regression models as baselines: a similarity proxy on typological similarity alone, and a generalized linear model on typological similarity, script, genealogical, and geographical distances.
We also investigate a control of equal model capacity which uses five non-typological descriptors: writing script, WALS family, WALS genus, geographic region, and Wikipedia resource bin.
A language is put into on of four resource bins, based on the amount of articles available in that language, or a separate bin if no articles available.
We use leave-one-language-out (LOLO) and leave-$k$-languages-out cross-validation, reporting held-out Pearson $\rho$ for ranking performance and $R^2$ for regression performance.
In addition, we use leave-one-family-out (LOFO) and leave-one-script-out (LOSO) as more general holdouts.
A fold-respecting permutation null, cluster bootstrap intervals by source language, and paired bootstrap tests of $\Delta\rho$ are employed for statistical tests.
Refer to Appendix~\ref{app:config} for details.
\paragraph{Resource-bias decomposition}
To assess how strongly resources and script shape transfer, we model $\mathrm{BTS} = T + b$, composed of typological signal $T$ and resource-and-script bias $b$, which we model as Wikipedia article counts (source and target language separate), script difference, and genealogical and geographic distance.
We fit a ridge regression to estimate $\hat b$ and then a random forest that fits the typology term $\hat T$ on the residual $\mathrm{BTS} - \hat b$.

\paragraph{Experiments}
Experiments are grouped as follows: E1--E4 probe RQ1 and E5 probes RQ2.
E1 examines whether typological features reconstruct measured transfer, by comparing held-out performance across the four model tiers and the matched-capacity metadata control against the permutation null.
E2 examines whether the reconstruction is carried by shared writing script rather than by typology, comparing cross-script, same-script, and Latin-only held-out pairs against a script-only directed baseline.
E3 examines the genealogical confound through grouped holdouts that leave an entire script or an entire family out of training at once.
E4 examines where the signal lives, sweeping the number of features from a single one to the full set and measuring the stability of the importance ranking across folds.
E5 examines which source language transfers best, and how a source's data availability shapes this result, by ranking candidate sources under four operationalizations of transfer that vary in how much resource bias they admit.
Full experimental configurations are in Appendix~\ref{app:config}.

\FloatBarrier
\section{Results}\label{sec:results}
\FloatBarrier
\paragraph{Reconstruction performance across model capacity (E1)}\looseness=-1
The reconstruction performance for different model setups is given in Table~\ref{tab:tiers}.
The predictive performance of the similarity proxy and ridge GLM baselines in both protocols does not meet the simple baseline threshold $\rho{=}0.225$ (Appendix~\ref{app:ablate}, Table~\ref{tab:extrabaselines}).
The untuned random forest passes this threshold and is able to empirically reconstruct the ATLAS transfer matrix.
Tuning the model using a hyperparameter sweep improves the performance of the model, highlighting the value of the typological features for transfer prediction.
A nested rerun of the sweep within each training fold scores $\rho{=}0.58$, the honest bound on tuning at $n{=}24$ (Appendix~\ref{app:ablate}). \\
\textbf{Ablations}
In harder protocols the results degrade but do not collapse, reaching $\rho{=}0.41$ in the worst case where both languages of each held-out pair are unseen (Appendix~\ref{app:ablate}, Table~\ref{tab:infer}).
In a matched-capacity setting, where metadata is compared against typology with 5 features each, the performance difference disappears, suggesting that the strength of typology lies in the breadth of the features (Appendix~\ref{app:ablate}, Table~\ref{tab:matched}).
Statistical inference supports the performance of the tuned model: the cluster bootstrap 95\% CI of $[0.51, 0.82]$ excludes the permutation null (mean $-0.007$) at $p{=}0.005$.
The ordering of feature sources survives metric and language-set changes (Tables~\ref{tab:sources} and~\ref{tab:shared22}).
Prediction heatmaps of all tiers are in Appendix~\ref{app:transfer}.
Interestingly, Grambank features alone outperform the Grambank + WALS union: $\rho{=}0.78$ vs. $0.705$ (Appendix~\ref{app:ablate}, Table~\ref{tab:sources}).
The union's degradation comes from the model fitting WALS patterns that hold only within the small training set and fail on held-out languages, rather than from column dilution or WALS feature quality (Appendix~\ref{app:ablate}, Table~\ref{tab:walsperm}).
For the final results, the Grambank + WALS union is adopted anyway to avoid post-hoc selection bias.
\begin{table}[h]
\centering\footnotesize
\begin{minipage}[t]{0.60\linewidth}
\centering
\setlength{\tabcolsep}{3pt}
\caption{LOLO and leave-2-languages-out predictive performance on ATLAS-24 (E1).}\label{tab:tiers}
\begin{tabular}{lccc}
\toprule
Model & LOLO $\rho$ & $R^2$ & L2LO $\rho$ \\
\midrule
similarity proxy                          & 0.037 & $-0.010$ & 0.029 \\
ridge GLM (sim/script/gene/geo)           & 0.162 & 0.014 & 0.154 \\
random forest (all features)               & 0.615 & 0.350 & 0.648 \\
\textbf{tuned RF (typology)}              & \textbf{0.705} & \textbf{0.492} & \textbf{0.702} \\
\midrule
metadata RF (matched-capacity, 5 fields)  & 0.581 & 0.264 & 0.611 \\
metadata RF (own-sweep tuned)            & 0.620 & 0.375 & 0.621 \\
single-feature ridges (script/family/wiki)& ${\le}0.17$ & ${\approx}0$ & ${\le}0.17$ \\
\bottomrule
\end{tabular}
\end{minipage}\hfill
\begin{minipage}[t]{0.36\linewidth}
\centering
\setlength{\tabcolsep}{3pt}
\caption{Cross-script robustness and grouped holdouts (E2, E3) on ATLAS-24.}\label{tab:ood}
\begin{tabular}{lc}
\toprule
Split & $\rho$ \\
\midrule
cross-script pairs                 & 0.728 \\
same-script pairs                  & 0.600 \\
Latin--Latin only                  & 0.509 \\
script-only model                  & 0.138 \\
\midrule
LOLO (leave-one-language-out)      & 0.705 \\
LOSO (leave-one-script-out)        & 0.776 \\
matched LOLO (LOSO pairs)          & 0.667 \\
LOFO (leave-one-family-out)        & 0.729 \\
matched LOLO (LOFO pairs)          & 0.671 \\
\bottomrule
\end{tabular}
\end{minipage}
\end{table}
\paragraph{Generalization across scripts and families (E2, E3)}\looseness=-1
Cross-script held-out pairs are predicted at least as well as same-script, at $\rho{=}0.728$ vs. $0.600$ (Table~\ref{tab:ood}), and the model stays well above the script-only baseline even on Latin-Latin pairs for source and target language, where script carries no information. 
Grouped holdout does not decrease performance either: LOSO and LOFO macro-$\rho$ exceed the matched LOLO baseline on the same held-out pairs, and the pair-weighted $\rho$ is unchanged (Appendix~\ref{app:ablate}, Table~\ref{tab:composition}).
A symmetric cross-check agrees: language pairs that share a script transfer more on average, but a script-only model cannot predict the directed scores and stays near chance (Appendix~\ref{app:ablate}, Tables~\ref{tab:infer} and~\ref{tab:symmetric}), highlighting that indeed script strongly correlates with transfer, but has no incremental directed power.
To conclude, holding out an entire group does not collapse performance ($p{<}0.01$ for paired bootstrap test), indicating that the signal is typological and not confounded by script or language family.
\paragraph{Where the signal lives (E4)}\looseness=-1
Typological signal for cross-lingual transfer is highly redundant rather than concentrated in a few decisive features (Table~\ref{tab:sweep}).
The top-$k$ sweep is flat up to $k{=}100$ ($\rho{=}0.57$--$0.65$), peaks at $k{=}200$ (0.742), and settles at $0.705$ for the full 387 feature set.
The top-25 importance ranking is almost completely unstable across folds, yet the single best feature ($k{=}1$) already reaches $\rho{=}0.622$, because typological features are correlated (Appendix~\ref{app:ablate}, Table~\ref{tab:infer}).
This suggests that hundreds of typological features encode one coarse signal, and any one of them recovers most of it.
\begin{table}[h]
\centering\footnotesize
\begin{minipage}[t]{0.29\linewidth}
\centering
\setlength{\tabcolsep}{3pt}
\caption{Top-$k$ feature sweep for Experiment E4 under Grambank+WALS; the leak-free column selects its top-$k$ within each fold.}\label{tab:sweep}
\begin{tabular}{lcc}
\toprule
top-$k$ & global-rank $\rho$ & leak-free $\rho$ \\
\midrule
1 & 0.622 & 0.467 \\
5 & 0.613 & 0.624 \\
10 & 0.573 & 0.692 \\
25 & 0.570 & 0.637 \\
50 & 0.647 & 0.656 \\
100 & 0.627 & 0.690 \\
200 & 0.742 & 0.734 \\
387 & 0.705 & 0.705 \\
\bottomrule
\end{tabular}
\end{minipage}\hfill
\begin{minipage}[t]{0.63\linewidth}
\centering
\setlength{\tabcolsep}{3pt}
\caption{``Best pre-training source'' over a fixed candidate set (Experiment E5); rows are not comparable across scales.}\label{tab:rq4}
\begin{tabular}{lll}
\toprule
Operationalization & \#1 source & \#2 source \\
\midrule
empirical (ATLAS, raw BTS mean)        & English ($+0.022$) & Hebrew ($-0.074$) \\
on-scale reweighted RF                 & English ($-0.005$) & Hebrew ($-0.070$) \\
OOD-generalizer (cross-family)         & English ($-0.011$) & Hebrew ($-0.065$) \\
\midrule
resource-adjusted residual             & Marathi ($+0.36$)  & Filipino ($+0.36$) \\
\bottomrule
\end{tabular}
\end{minipage}
\end{table}
\paragraph{The best pre-training source (E5)}\looseness=-1
Table~\ref{tab:rq4} lists the top-scoring languages per measure, along with their transfer scores.
On every BTS-scale operationalization, English ranks first in average transfer score across all languages.
The resource-adjusted residual changes this: Marathi and Filipino top the ranking instead.
This reversal shows that raw rankings are highly sensitive to the resource-and-script bias term, which typology never observes.
The flip itself is statistically robust: Marathi's mean residual exceeds English's by $0.24$ ($p{<}0.001$), holds under a leak-free refit of the bias layer, and clears a permutation null (Appendix~\ref{app:ablate}, Table~\ref{tab:infer}).
Across 200 reruns, the specific top language in the resource-adjusted ranking varies, so the robust finding is the flip away from English, rather than any particular ranking (Appendix~\ref{app:ablate}).

\section{Conclusion, Discussion and Limitations}\label{sec:concl}
\subsection{Conclusion and Discussion}
\looseness=-1
We show that a typology-only predictor reconstructs ATLAS transfer, scoring above a non-typological control under leave-one-language, leave-one-script, and leave-one-family holdouts.
The signal it uses is redundant rather than localized: hundreds of correlated features encode one coarse dimension, and a single feature recovers most of it.
Furthermore, we show that best-source rankings built from resource statistics inherit a resource-and-script bias; typological features never observe these resources, so they offer an independent lens on the same question.

Our model requires no new LLM training runs.
\looseness=-1
Given an existing transfer matrix to fit on, scoring any ground-truth language costs only a prediction of a random forest, which makes typology a promising candidate for prioritizing source languages, or for proposing data mixtures to test on low compute budget.
Further assessing the quality of such data mixtures requires more in-depth downstream analyses, which we leave for future work.

Beyond our results, the distributed signal also calls on linguistic theory: transfer emerges from many small structural alignments rather than any single feature family.

The bias decomposition warns that a source can look excellent merely because of resource abundance, raising the question whether current high-resource languages are the theoretical optimal training sources, especially in light of future dataset collection efforts.
\subsection{Limitations}
\paragraph{Ground truth and compute}
Our results rest entirely on the ATLAS matrix~\cite{ICLR2026_35c26a81}, which mixes measured BTS with FAS-estimated scores, so we model a noisy target whose confounds propagate to our findings.
Compute constraints confine us to ATLAS's published results, and strong claims about pre-training require verification at scale.
\paragraph{Scale and sampling}
All source rankings are 24-language extrapolations on the non-random ATLAS--Grambank overlap, which drops strong European sources such as Spanish and German, and the unseen-language $\rho{\approx}0.41$ is our generalization estimate.
\paragraph{Residual adjustment is a diagnostic}
The bias layer's weights describe ATLAS-24 rather than a stable law.
The decomposition depends on this choice of five covariates without causal identification, so the off-scale residual $T$ only exposes the ranking's bias-sensitivity (Appendix~\ref{app:adds}).
Residuals are mean-zero by construction and unitless with respect to BTS, so they support only relative comparisons. A script-by-genus reweighted random forest on raw BTS scores keeps predictions on-scale and comparable.

\FloatBarrier
\begin{ack}
This research was supported by the OpenEuroLLM project, co-funded by the Digital Europe Programme under GA no. 101195233.
\end{ack}

\FloatBarrier
\bibliographystyle{abbrvnat}
\bibliography{refs}


\appendix

\FloatBarrier
\section{Experimental configuration}\label{app:config}
\FloatBarrier
\paragraph{Notation}
Formally, the input is $[M_s, M_t, \Delta]$ with $M_s, M_t \in \{0,1\}^{n_\text{codes}}$ the one-hot per-feature profiles ($n_{\text{codes}} = \sum_{f \in F} |C_f|$, with $C_f$ the codes of feature $f$) and $\Delta \in \{0,1\}^{n_\text{feat}}$ a per-feature disagreement block ($n_{\text{feat}} = |F|$).
\paragraph{Protocols, metrics, and controls in plain language}
Every experiment trains a model on some language pairs and then scores it on pairs it never trained on, and the protocols below differ only in how the pairs are split.
In leave-one-language-out, every pair touching one language is hidden, the model trains on the remaining pairs and predicts the hidden ones, and each of the 24 languages takes a turn as the hidden language.
Leave-$k$-languages-out hides $k$ languages at once, so a test pair can involve languages that never appear in training, which is the harder setting.
Pearson $\rho$ checks whether the model orders pairs as ATLAS does: $1$ means a perfect ordering, $0$ means the predicted ordering carries no information about the true one, and negative values mean systematically reversed orderings.
$R^2$ asks the stronger question of whether the predicted values are numerically close: it compares the model's squared error against that of a baseline that always predicts the average BTS, so $1$ is perfect, $0$ matches the baseline, and negative is worse than it.
This is the quantity also known as explained variance, the share of the variation in BTS that the predictions account for.
The permutation null shuffles the training scores within each fold, reruns the procedure, and reports how often such chance data score as well as the real data.
Because pairs that share a language are not independent, the cluster bootstrap resamples whole languages with replacement, and each drawn language brings all of its pairs.
Repeating the evaluation on each resample expresses how much the reported score could plausibly vary.
The matched-capacity control is a random forest of identical size and tuning budget that receives five non-typological descriptors instead of typological features.
A gap in its favor therefore reflects the typological features, rather than model capacity or tuning.
\begin{table}[h]
\centering\small
\caption{Shared setup underlying all experiments in Table~\ref{tab:config}.
All typology experiments use the Grambank+WALS feature source and the one-decimal ATLAS target at $n{=}24$, which is 552 directed pairs, unless a row of Table~\ref{tab:config} states a deviation.}\label{tab:config_shared}
\begin{tabular}{@{}p{1.25in}p{4.00in}@{}}
\toprule
Component & Exact configuration \\
\midrule
Empirical target & the directed BTS transfer matrix digitized from Figure~C.2 of the ATLAS paper, read at one decimal \\
Typological features & Grambank with 195 features and WALS with 192, keyed by Glottocode; the union carries 387 features \\
Modeling set & ATLAS-24: the 24 ATLAS languages with Grambank coverage of at least $0.70$; the floor is Grambank-only, so WALS adds feature columns but no languages; all $24 \times 23$ directed pairs with $s \neq t$, which is 552 \\
Input representation & $[M_s, M_t, \Delta]$, the one-hot code profiles of source and target plus a per-feature disagreement block, as defined above \\
Main model & random forest with 200 trees, \texttt{random\_state}$=42$, and \texttt{max\_features}$=0.5$; under WALS alone \texttt{max\_features} is \texttt{log2}; all remaining parameters at scikit-learn defaults \\
Ridge models & $\alpha{=}1.0$ throughout: the similarity proxy, the GLM on similarity, script, genealogical, and geographic distances, the script-only model, and the feature-importance ranker; the bias ridge standardizes its inputs first \\
Cross-validation & leave-one-language-out with 24 folds; a pair is held out exactly when either endpoint is the held-out language; each fold refits the model; held-out predictions are pooled over folds, and a pair predicted in two folds resolves to the fold of its alphabetically later language \\
Metrics & held-out Pearson $\rho$ and true $R^2 = 1 - \text{SS}_\text{res}/\text{SS}_\text{tot}$ \\
Seeds and records & seed 42 throughout, with seeds 43 and 44 for seed stability; every result CSV carries a \texttt{\_meta.json} holding the resolved configuration and the git commit; locked environment: Python 3.12, scikit-learn 1.9, NumPy 2.5, pandas 3.0, SciPy 1.18 \\
\bottomrule
\end{tabular}
\end{table}
\begin{longtable}{@{}p{1.25in}p{4.00in}@{}}
\caption{Exact configuration of the five experiments E1 to E5 and their sub-analyses.
Each experiment addresses the research question named in its group header, with E1 to E4 probing RQ1 and E5 probing RQ2.
Settings inherited from Table~\ref{tab:config_shared} are not repeated.
Repository scripts preserve their own identifiers, so the experiment labels E1 to E5 are assigned here for readability.}\label{tab:config}\\
\toprule
Analysis & Exact configuration \\
\midrule
\endfirsthead
\caption[]{Exact configuration of Experiments E1 to E5, continued.}\\
\toprule
Analysis & Exact configuration \\
\midrule
\endhead
\bottomrule
\endfoot
\multicolumn{2}{@{}l}{\textit{E1 (RQ1): held-out performance and capacity}} \\
Tier comparison & four tiers under identical folds: the similarity-proxy ridge, the GLM ridge, the 200-tree RF at scikit-learn defaults, and the tuned RF; metadata single-feature ridges for script, family, and Wikipedia and their combination, each with within-fold standardization on its own coverage-restricted universe: script and Wikipedia cover 24 languages, the WALS family map covers 22 \\
Distance baselines & five non-typological distance baselines under the identical folds and metrics, each a single-column ridge except where noted, with within-fold standardization on the full 24-language universe: the lang2vec cosine over the concatenated imputed syntax, phonology, and inventory vectors; the URIEL design with six columns, genetic from the language-family vector, geographic as the great-circle distance on Glottolog coordinates normalized by half the circumference, syntactic, phonological, and inventory as one minus the cosine of the respective vectors, and featural as their mean; and three Jaccard overlaps over per-language Wikipedia lead-paragraph corpora of 2{,}000 random articles per edition with a committed sha256 manifest, namely byte-level BPE vocabularies of 8{,}000 entries, pooled character 1-to-3-gram sets over the 5{,}000 most frequent word types, and the 10{,}000 most frequent word types; lang2vec 1.1.2 imputed feature sets cover all 24 languages, and the learned embeddings are unusable because seven of the 24 languages are absent and the embedding file does not load under NumPy 2 \\
Leave-$k$-languages-out & $k \in \{1,2,3\}$; $k{=}1$ is the deterministic LOLO partition; for $k \geq 2$ the tuned RF averages over 50 random disjoint partitions of the 24 languages and the remaining tiers over 25; a pair is held out when either endpoint is in the held-out group; the reported $\rho$ is the partition mean and the interval is the partition spread; seed 42 \\
Permutation null & fold-respecting: within each fold only the training labels are permuted and the held-out labels are untouched; $B{=}200$; $p = (1 + \#\{\text{null} \geq \text{observed}\}) / (B{+}1)$; seed 42 \\
Cluster bootstrap & 95\% intervals by resampling the 24 source languages with replacement, $B{=}2000$, seed 42; paired bootstrap of $\Delta\rho$ against the script-only ridge and against the GLM on shared pairs; a language-level variant resamples the 24 languages and reweights each pair by the product of how often its two endpoints were drawn \\
Feature sources & Grambank with 195 features and Grambank+WALS with 387 run at $n{=}24$; WALS runs at $n{=}22$ because WALS lacks Glottocode mappings for two of the 24 languages; each source carries its own swept \texttt{max\_features}, which is 0.5, 0.5, and \texttt{log2} in the order above \\
Shared-language sources & all three sources refit on the 22 ATLAS-24 languages covered by the WALS mapping, 462 directed pairs; per-source \texttt{max\_features} mini-sweep over the seven grid values; identical LOLO folds; seed 42 \\
WALS-block permutation & the WALS block of the wide matrix permuted across the 24 languages, preserving each feature's marginal levels and missingness; ten permutations, each under the seven-value \texttt{max\_features} mini-sweep on the production LOLO folds at $n{=}24$; forest seed 42, permutation seeds 1000 to 1009 \\
Coverage-floor sweep & the per-language Grambank coverage floor lowered from $0.70$ to $0.60$, $0.50$, $0.40$, and $0.00$, admitting one language per step, namely Telugu, Thai, Indonesian, and Central Kurdish; every tier refit at each floor under the production configuration and the identical LOLO protocol; the tuned RF is additionally scored under the pure-unseen-language protocol at each floor; seed 42 \\
Digitization-noise robustness & 100 refits of the tuned RF on targets perturbed by independent $U(-0.05, 0.05)$ noise per cell, forest seed fixed at 42, scored on the clean target \\
Pure-unseen-language protocol & repeated grouped 5-fold over languages with five repeats; test pairs are those whose both languages are held out, so training pairs never touch a held-out language; seed 42, with three seeds in the refinement stage \\
Matched-capacity control & a non-typological RF over the one-hot per-language metadata profile: script with 9 levels, WALS family, WALS genus, a $k$-means geographic region with 6 clusters, and a Wikipedia article-count quartile bin; missing values get an explicit unknown level, which favors metadata and is the conservative direction; the same pair design, 200-tree RF, and LOLO folds as the typology model; the budget-matched row averages LOLO $\rho$ over 20 random 5-feature typology subsets drawn from the top 15 of the global importance ranking, seed 42 \\
Matched-tuning control & the metadata RF swept over the same curated grid as the typology model; all 168 configurations screened under LOLO at 200 trees, seed 42; top 8 refined over seeds 42 to 46; the winner by true $R^2$ is \texttt{max\_features} \texttt{sqrt} with \texttt{max\_depth} 10, \texttt{min\_samples\_leaf} 1, and \texttt{max\_samples} 1.0; leave-2-languages-out over 50 random partitions; paired cluster bootstrap of $\Delta\rho$ against the tuned typology model with source-language blocks, $B{=}2000$, seed 42 \\
Hyperparameter sweep & curated grid over \texttt{max\_features} $\in \{0.2, 0.3, 0.5, 0.7, 1.0, \texttt{sqrt}, \texttt{log2}\}$, \texttt{max\_depth} $\in \{\texttt{None}, 6, 10, 16\}$, \texttt{min\_samples\_leaf} $\in \{1, 2, 5\}$, and \texttt{max\_samples} $\in \{0.6, 1.0\}$; screening under LOLO at 200 trees and one seed; refinement of the top 8 configurations plus the scikit-learn-default and previously tuned references under both protocols at 200 and 500 trees, with five LOLO seeds and three unseen-language seeds; the metric of record is true $R^2$ with Pearson as tie-break; the per-source winner maximizes the combined rank of LOLO and unseen-language $R^2$; a fully nested rerun of this selection inside each outer training fold (inner LOLO plus inner grouped holdout over the whole grid) scores $\rho{=}0.580$ for Grambank+WALS with 21 of 24 distinct per-fold winners, bounding what the tuning step contributes honestly \\
\midrule
\multicolumn{2}{@{}l}{\textit{E2 (RQ1): within-script typology isolation}} \\
Script stratification & pooled LOLO predictions split into same-script, cross-script, and Latin--Latin pairs; four models: the script-only ridge, the GLM with the script column, the GLM without the script column, and the tuned RF; run under Grambank+WALS and under Grambank alone; subgroup 95\% intervals by cluster bootstrap, $B{=}2000$ \\
\midrule
\multicolumn{2}{@{}l}{\textit{E3 (RQ1): out-of-distribution taxonomy}} \\
Grouped holdout & leave-one-script-out with groups by writing script and leave-one-family-out with groups by WALS major family; the fold for group $g$ holds out every pair touching a language of $g$; per-group $\rho$ is scored when the group has at least three held-out pairs; the macro value is the unweighted mean over groups; 95\% intervals by one-sample bootstrap over held-out groups, $B{=}2000$ \\
Composition-matched baseline & pooled LOLO predictions restricted to the held-out pairs of each script or family group, macro-averaged and pair-weighted, compared against grouped holdout on the same groups \\
Symmetric cross-check & per-factor Mantel $r$ and MRM with $B{=}999$ permutations of the response-matrix labels on the 276 upper-triangle cells; the response is symmetrized BTS, the mean of the two directions, treated as a distance; predictors are min--max-normalized typological, genealogical, geographic, and script distances; partial Mantel is deliberately not used because of its Type-I error inflation \\
\midrule
\multicolumn{2}{@{}l}{\textit{E4 (RQ1): where the signal lives}} \\
Feature ranking & ridge permutation importance with $\alpha{=}1$ and seed 42; the importance of a feature is the drop in held-out $\rho$ when its one-hot columns and disagreement bit are permuted within each fold's training set; the global ranking pools this over folds, and the stability analysis keeps the 24 per-fold rankings \\
Top-$k$ sweep & $k \in \{1, 5, 10, 25, 50, 100, 200\}$ plus the full 387; features ordered by the global ranking; the same tuned RF; the ranking is scored on the evaluation folds, so entries with $k < 387$ are optimistically biased \\
Leak-free selection & each fold takes its own top-$k$ by inner-LOLO importance computed entirely within that fold's training pairs, so the held-out language touches neither the fit nor the ranking; the same RF; seed stability over seeds 42, 43, and 44 \\
Importance stability & mean pairwise Spearman correlation over the 24 per-fold importance vectors and mean pairwise top-25 Jaccard overlap; 95\% interval by bootstrap over fold pairs, $B{=}2000$ \\
\midrule
\multicolumn{2}{@{}l}{\textit{E5 (RQ2): the best pre-training source}} \\
Operationalizations & the candidate set and the target set are both the full ATLAS-24; empirical is the mean raw BTS per source over targets; the resource-adjusted residual is the mean residual $T$, with an on-scale variant $\hat T + \hat B$; centrality is the mean typological similarity to the other 23 languages on the Grambank+WALS features; the OOD-generalizer is the mean raw BTS to targets from a different WALS major family; the on-scale reweighted model is the 200-tree RF on raw BTS with sample weight $1/|\text{block}|$ per unordered script-by-genus block; the top 10 sources are reported per operationalization, and the full 24 in Table~\ref{tab:e5_fullrank} \\
Bias layer & ridge with $\alpha{=}1$ on standardized script difference, genealogical distance, min--max-normalized geographic distance, and $\log1p$ Wikipedia article counts of source and target, fitted in-sample on all 552 pairs; the residual $T = \mathrm{BTS} - \hat B$ is mean-zero by construction; robustness is probed by refitting on all 31 non-empty descriptor subsets, Table~\ref{tab:bias_ablation} \\
Leak-free bias layer & $\hat B$ refit out-of-fold under the LOLO folds, each pair's bias predicted by a ridge that never saw its held-out language, and the residual ranking recomputed as the raw mean residual and through the production rf$_T$ \\
Residual permutation null & the full operationalization, bias ridge, residual, rf$_T$, and per-source ranking, rerun on permuted BTS labels, $B{=}200$, seed 42, recording the top-source margin over the median source and the gap over English per replicate \\
Residual gap test & the difference of mean residuals between two sources is bootstrapped over targets, resampling targets and pairing both sources on the same resample, $B{=}2000$ \\
\end{longtable}

\paragraph{Compute and provenance}
All experiments ran on a single CPU of a local workstation, and no GPU, cluster, or cloud resources were used.
The models are scikit-learn random forests and ridge regressions on 552 directed pairs, so each configuration fits in seconds at this scale.
Grambank v1.0 and WALS Online are distributed under CC-BY 4.0, lang2vec under CC-BY-SA 4.0, scikit-learn under the BSD 3-Clause license, and Wikipedia text under CC-BY-SA 4.0.
The ATLAS paper is published at ICLR 2026, and we digitize the values of its Figure~C.2 for analysis only.
Our released code is MIT-licensed.

\FloatBarrier
\begingroup
\footnotesize
\begin{longtable}{@{}lp{0.40\linewidth}lp{0.40\linewidth}@{}}
\caption{The 387 typological features used in this work, in natural ID order within each source: 195 Grambank parameters and 192 WALS features.}\label{tab:features}\\
\toprule
ID & Feature & ID & Feature \\
\midrule
\endfirsthead
\multicolumn{4}{@{}l}{\emph{(continued)}}\\
\midrule
ID & Feature & ID & Feature \\
\midrule
\endhead
\multicolumn{4}{@{}l}{\emph{Grambank} (195 features)}\\[2pt]
GB020 & Are there definite or specific articles? & GB021 & Do indefinite nominals commonly have indefinite articles? \\
GB022 & Are there prenominal articles? & GB023 & Are there postnominal articles? \\
GB024 & What is the order of numeral and noun in the NP? & GB025 & What is the order of adnominal demonstrative and noun? \\
GB026 & Can adnominal property words occur discontinuously? & GB027 & Are nominal conjunction and comitative expressed by different elements? \\
GB028 & Is there a distinction between inclusive and exclusive? & GB030 & Is there a gender distinction in independent 3rd person pronouns? \\
GB031 & Is there a dual or unit augmented form (in addition to plural or augmented) for all person categories in the pronoun system? & GB035 & Are there three or more distance contrasts in demonstratives? \\
GB036 & Do demonstratives show an elevation distinction? & GB037 & Do demonstratives show a visible-nonvisible distinction? \\
GB038 & Are there demonstrative classifiers? & GB039 & Is there nonphonological allomorphy of noun number markers? \\
GB041 & Are there several nouns (more than three) which are suppletive for number? & GB042 & Is there productive overt morphological singular marking on nouns? \\
GB043 & Is there productive morphological dual marking on nouns? & GB044 & Is there productive morphological plural marking on nouns? \\
GB046 & Is there an associative plural marker for nouns? & GB047 & Is there a productive morphological pattern for deriving an action/state noun from a verb? \\
GB048 & Is there a productive morphological pattern for deriving an agent noun from a verb? & GB049 & Is there a productive morphological pattern for deriving an object noun from a verb? \\
GB051 & Is there a gender/noun class system where sex is a factor in class assignment? & GB052 & Is there a gender/noun class system where shape is a factor in class assignment? \\
GB053 & Is there a gender/noun class system where animacy is a factor in class assignment? & GB054 & Is there a gender/noun class system where plant status is a factor in class assignment? \\
GB057 & Are there numeral classifiers? & GB058 & Are there possessive classifiers? \\
GB059 & Is the adnominal possessive construction different for alienable and inalienable nouns? & GB065 & What is the pragmatically unmarked order of adnominal possessor noun and possessed noun? \\
GB068 & Do core adjectives (defined semantically as property concepts such as value, shape, age, dimension) act like verbs in predicative position? & GB069 & Do core adjectives (defined semantically as property concepts; value, shape, age, dimension) used attributively require the same morphological treatment as verbs? \\
GB070 & Are there morphological cases for non-pronominal core arguments (i.e. S/A/P)? & GB071 & Are there morphological cases for pronominal core arguments (i.e. S/A/P)? \\
GB072 & Are there morphological cases for oblique non-pronominal NPs (i.e. not S/A/P)? & GB073 & Are there morphological cases for independent oblique personal pronominal arguments (i.e. not S/A/P)? \\
GB074 & Are there prepositions? & GB075 & Are there postpositions? \\
GB079 & Do verbs have prefixes/proclitics, other than those that only mark A, S or P (do include portmanteau: A \& S + TAM)? & GB080 & Do verbs have suffixes/enclitics, other than those that only mark A, S or P (do include portmanteau: A \& S + TAM)? \\
GB081 & Is there productive infixation in verbs? & GB082 & Is there overt morphological marking of present tense on verbs? \\
GB083 & Is there overt morphological marking on the verb dedicated to past tense? & GB084 & Is there overt morphological marking on the verb dedicated to future tense? \\
GB086 & Is a morphological distinction between perfective and imperfective aspect available on verbs? & GB089 & Can the S argument be indexed by a suffix/enclitic on the verb in the simple main clause? \\
GB090 & Can the S argument be indexed by a prefix/proclitic on the verb in the simple main clause? & GB091 & Can the A argument be indexed by a suffix/enclitic on the verb in the simple main clause? \\
GB092 & Can the A argument be indexed by a prefix/proclitic on the verb in the simple main clause? & GB093 & Can the P argument be indexed by a suffix/enclitic on the verb in the simple main clause? \\
GB094 & Can the P argument be indexed by a prefix/proclitic on the verb in the simple main clause? & GB095 & Are variations in marking strategies of core participants based on TAM distinctions? \\
GB096 & Are variations in marking strategies of core participants based on verb classes? & GB098 & Are variations in marking strategies of core participants based on person distinctions? \\
GB099 & Can verb stems alter according to the person of a core participant? & GB103 & Is there a benefactive applicative marker on the verb (including indexing)? \\
GB104 & Is there an instrumental applicative marker on the verb (including indexing)? & GB105 & Can the recipient in a ditransitive construction be marked like the monotransitive patient? \\
GB107 & Can standard negation be marked by an affix, clitic or modification of the verb? & GB108 & Is there directional or locative morphological marking on verbs? \\
GB109 & Is there verb suppletion for participant number? & GB110 & Is there verb suppletion for tense or aspect? \\
GB111 & Are there conjugation classes? & GB113 & Are there verbal affixes or clitics that turn intransitive verbs into transitive ones? \\
GB114 & Is there a phonologically bound reflexive marker on the verb? & GB115 & Is there a phonologically bound reciprocal marker on the verb? \\
GB116 & Do verbs classify the shape, size or consistency of absolutive arguments by means of incorporated nouns, verbal affixes or suppletive verb stems? & GB117 & Is there a copula for predicate nominals? \\
GB118 & Are there serial verb constructions? & GB119 & Can mood be marked by an inflecting word ("auxiliary verb")? \\
GB120 & Can aspect be marked by an inflecting word ("auxiliary verb")? & GB121 & Can tense be marked by an inflecting word ("auxiliary verb")? \\
GB122 & Is verb compounding a regular process? & GB123 & Are there verb-adjunct (aka light-verb) constructions? \\
GB124 & Is incorporation of nouns into verbs a productive intransitivizing process? & GB126 & Is there an existential verb? \\
GB127 & Are different posture verbs used obligatorily depending on an inanimate locatum's shape or position (e.g. 'to lie' vs. 'to stand')? & GB129 & Is there a notably small number, i.e. about 100 or less, of verb roots in the language? \\
GB130 & What is the pragmatically unmarked order of S and V in intransitive clauses? & GB131 & Is a pragmatically unmarked constituent order verb-initial for transitive clauses? \\
GB132 & Is a pragmatically unmarked constituent order verb-medial for transitive clauses? & GB133 & Is a pragmatically unmarked constituent order verb-final for transitive clauses? \\
GB134 & Is the order of constituents the same in main and subordinate clauses? & GB135 & Do clausal objects usually occur in the same position as nominal objects? \\
GB136 & Is the order of core argument (i.e. S/A/P) constituents fixed? & GB137 & Can standard negation be marked clause-finally? \\
GB138 & Can standard negation be marked clause-initially? & GB139 & Is there a difference between imperative (prohibitive) and declarative negation constructions? \\
GB140 & Is verbal predication marked by the same negator as all of the following types of predication: locational, existential and nominal? & GB146 & Is there a morpho-syntactic distinction between predicates expressing controlled versus uncontrolled events or states? \\
GB147 & Is there a morphological passive marked on the lexical verb? & GB148 & Is there a morphological antipassive marked on the lexical verb? \\
GB149 & Is there a morphologically marked inverse on verbs? & GB150 & Is there clause chaining? \\
GB151 & Is there an overt verb marker dedicated to signalling coreference or noncoreference between the subject of one clause and an argument of an adjacent clause ("switch reference")? & GB152 & Is there a morphologically marked distinction between simultaneous and sequential clauses? \\
GB155 & Are causatives formed by affixes or clitics on verbs? & GB156 & Is there a causative construction involving an element that is unmistakably grammaticalized from a verb for 'to say'? \\
GB158 & Are verbs reduplicated? & GB159 & Are nouns reduplicated? \\
GB160 & Are elements apart from verbs or nouns reduplicated? & GB165 & Is there productive morphological trial marking on nouns? \\
GB166 & Is there productive morphological paucal marking on nouns? & GB167 & Is there a logophoric pronoun? \\
GB170 & Can an adnominal property word agree with the noun in gender/noun class? & GB171 & Can an adnominal demonstrative agree with the noun in gender/noun class? \\
GB172 & Can an article agree with the noun in gender/noun class? & GB177 & Can the verb carry a marker of animacy of argument, unrelated to any gender/noun class of the argument visible in the NP domain? \\
GB184 & Can an adnominal property word agree with the noun in number? & GB185 & Can an adnominal demonstrative agree with the noun in number? \\
GB186 & Can an article agree with the noun in number? & GB187 & Is there any productive diminutive marking on the noun (exclude marking by system of nominal classification only)? \\
GB188 & Is there any productive augmentative marking on the noun (exclude marking by system of nominal classification only)? & GB192 & Is there a gender system where a noun's phonological properties are a factor in class assignment? \\
GB193 & What is the order of adnominal property word and noun? & GB196 & Is there a male/female distinction in 2nd person independent pronouns? \\
GB197 & Is there a male/female distinction in 1st person independent pronouns? & GB198 & Can an adnominal numeral agree with the noun in gender/noun class? \\
GB203 & What is the order of the adnominal collective universal quantifier ('all') and the noun? & GB204 & Do collective ('all') and distributive ('every') universal quantifiers differ in their forms or their syntactic positions? \\
GB250 & Can predicative possession be expressed with a transitive 'habeo' verb? & GB252 & Can predicative possession be expressed with an S-like possessum and a locative-coded possessor? \\
GB253 & Can predicative possession be expressed with an S-like possessum and a dative-coded possessor? & GB254 & Can predicative possession be expressed with an S-like possessum and a possessor that is coded like an adnominal possessor? \\
GB256 & Can predicative possession be expressed with an S-like possessor and a possessum that is coded like a comitative argument? & GB257 & Can polar interrogation be marked by intonation only? \\
GB260 & Can polar interrogation be indicated by a special word order? & GB262 & Is there a clause-initial polar interrogative particle? \\
GB263 & Is there a clause-final polar interrogative particle? & GB264 & Is there a polar interrogative particle that most commonly occurs neither clause-initially nor clause-finally? \\
GB265 & Is there a comparative construction that includes a form that elsewhere means 'surpass, exceed'? & GB266 & Is there a comparative construction that employs a marker of the standard which elsewhere has a locational meaning? \\
GB270 & Can comparatives be expressed using two conjoined clauses? & GB273 & Is there a comparative construction with a standard marker that elsewhere has neither a locational meaning nor a 'surpass/exceed' meaning? \\
GB275 & Is there a bound comparative degree marker on the property word in a comparative construction? & GB276 & Is there a non-bound comparative degree marker modifying the property word in a comparative construction? \\
GB285 & Can polar interrogation be marked by a question particle and verbal morphology? & GB286 & Can polar interrogation be indicated by overt verbal morphology only? \\
GB291 & Can polar interrogation be marked by tone? & GB296 & Is there a phonologically or morphosyntactically definable class of ideophones that includes ideophones depicting imagery beyond sound? \\
GB297 & Can polar interrogation be indicated by a V-not-V construction? & GB298 & Can standard negation be marked by an inflecting word ("auxiliary verb")? \\
GB299 & Can standard negation be marked by a non-inflecting word ("auxiliary particle")? & GB300 & Does the verb for 'give' have suppletive verb forms? \\
GB301 & Is there an inclusory construction? & GB302 & Is there a phonologically free passive marker ("particle" or "auxiliary")? \\
GB303 & Is there a phonologically free antipassive marker ("particle" or "auxiliary")? & GB304 & Can the agent be expressed overtly in a passive clause? \\
GB305 & Is there a phonologically independent reflexive pronoun? & GB306 & Is there a phonologically independent non-bipartite reciprocal pronoun? \\
GB309 & Are there multiple past or multiple future tenses, distinguishing distance from Time of Reference? & GB312 & Is there overt morphological marking on the verb dedicated to mood? \\
GB313 & Are there special adnominal possessive pronouns that are not formed by an otherwise regular process? & GB314 & Can augmentative meaning be expressed productively by a shift of gender/noun class? \\
GB315 & Can diminutive meaning be expressed productively by a shift of gender/noun class? & GB316 & Is singular number regularly marked in the noun phrase by a dedicated phonologically free element? \\
GB317 & Is dual number regularly marked in the noun phrase by a dedicated phonologically free element? & GB318 & Is plural number regularly marked in the noun phrase by a dedicated phonologically free element? \\
GB319 & Is trial number regularly marked in the noun phrase by a dedicated phonologically free element? & GB320 & Is paucal number regularly marked in the noun phrase by a dedicated phonologically free element? \\
GB321 & Is there a large class of nouns whose gender/noun class is not phonologically or semantically predictable? & GB322 & Is there grammatical marking of direct evidence (perceived with the senses)? \\
GB323 & Is there grammatical marking of indirect evidence (hearsay, inference, etc.)? & GB324 & Is there an interrogative verb for content interrogatives (who?, what?, etc.)? \\
GB325 & Is there a count/mass distinction in interrogative quantifiers? & GB326 & Do (nominal) content interrogatives normally or frequently occur in situ? \\
GB327 & Can the relative clause follow the noun? & GB328 & Can the relative clause precede the noun? \\
GB329 & Are there internally-headed relative clauses? & GB330 & Are there correlative relative clauses? \\
GB331 & Are there non-adjacent relative clauses? & GB333 & Is there a decimal numeral system? \\
GB334 & Is there synchronic evidence for any element of a quinary numeral system? & GB335 & Is there synchronic evidence for any element of a vigesimal numeral system? \\
GB336 & Is there a body-part tallying system? & GB400 & Are all person categories neutralized in some voice, tense, aspect, mood and/or negation? \\
GB401 & Is there a class of patient-labile verbs? & GB402 & Does the verb for 'see' have suppletive verb forms? \\
GB403 & Does the verb for 'come' have suppletive verb forms? & GB408 & Is there any accusative alignment of flagging? \\
GB409 & Is there any ergative alignment of flagging? & GB410 & Is there any neutral alignment of flagging? \\
GB415 & Is there a politeness distinction in 2nd person forms? & GB421 & Is there a preposed complementizer in complements of verbs of thinking and/or knowing? \\
GB422 & Is there a postposed complementizer in complements of verbs of thinking and/or knowing? & GB430 & Can adnominal possession be marked by a prefix on the possessor? \\
GB431 & Can adnominal possession be marked by a prefix on the possessed noun? & GB432 & Can adnominal possession be marked by a suffix on the possessor? \\
GB433 & Can adnominal possession be marked by a suffix on the possessed noun? & GB519 & Can mood be marked by a non-inflecting word ("auxiliary particle")? \\
GB520 & Can aspect be marked by a non-inflecting word ("auxiliary particle")? & GB521 & Can tense be marked by a non-inflecting word ("auxiliary particle")? \\
GB522 & Can the S or A argument be omitted from a pragmatically unmarked clause when the referent is inferrable from context ("pro-drop" or "null anaphora")? &  &  \\
\midrule
\multicolumn{4}{@{}l}{\emph{WALS} (192 features)}\\[2pt]
1A & Consonant Inventories & 2A & Vowel Quality Inventories \\
3A & Consonant-Vowel Ratio & 4A & Voicing in Plosives and Fricatives \\
5A & Voicing and Gaps in Plosive Systems & 6A & Uvular Consonants \\
7A & Glottalized Consonants & 8A & Lateral Consonants \\
9A & The Velar Nasal & 10A & Vowel Nasalization \\
10B & Nasal Vowels in West Africa & 11A & Front Rounded Vowels \\
12A & Syllable Structure & 13A & Tone \\
14A & Fixed Stress Locations & 15A & Weight-Sensitive Stress \\
16A & Weight Factors in Weight-Sensitive Stress Systems & 17A & Rhythm Types \\
18A & Absence of Common Consonants & 19A & Presence of Uncommon Consonants \\
20A & Fusion of Selected Inflectional Formatives & 21A & Exponence of Selected Inflectional Formatives \\
21B & Exponence of Tense-Aspect-Mood Inflection & 22A & Inflectional Synthesis of the Verb \\
23A & Locus of Marking in the Clause & 24A & Locus of Marking in Possessive Noun Phrases \\
25A & Locus of Marking: Whole-language Typology & 25B & Zero Marking of A and P Arguments \\
26A & Prefixing vs. Suffixing in Inflectional Morphology & 27A & Reduplication \\
28A & Case Syncretism & 29A & Syncretism in Verbal Person/Number Marking \\
30A & Number of Genders & 31A & Sex-based and Non-sex-based Gender Systems \\
32A & Systems of Gender Assignment & 33A & Coding of Nominal Plurality \\
34A & Occurrence of Nominal Plurality & 35A & Plurality in Independent Personal Pronouns \\
36A & The Associative Plural & 37A & Definite Articles \\
38A & Indefinite Articles & 39A & Inclusive/Exclusive Distinction in Independent Pronouns \\
39B & Inclusive/Exclusive Forms in Pama-Nyungan & 40A & Inclusive/Exclusive Distinction in Verbal Inflection \\
41A & Distance Contrasts in Demonstratives & 42A & Pronominal and Adnominal Demonstratives \\
43A & Third Person Pronouns and Demonstratives & 44A & Gender Distinctions in Independent Personal Pronouns \\
45A & Politeness Distinctions in Pronouns & 46A & Indefinite Pronouns \\
47A & Intensifiers and Reflexive Pronouns & 48A & Person Marking on Adpositions \\
49A & Number of Cases & 50A & Asymmetrical Case-Marking \\
51A & Position of Case Affixes & 52A & Comitatives and Instrumentals \\
53A & Ordinal Numerals & 54A & Distributive Numerals \\
55A & Numeral Classifiers & 56A & Conjunctions and Universal Quantifiers \\
57A & Position of Pronominal Possessive Affixes & 58A & Obligatory Possessive Inflection \\
58B & Number of Possessive Nouns & 59A & Possessive Classification \\
60A & Genitives, Adjectives and Relative Clauses & 61A & Adjectives without Nouns \\
62A & Action Nominal Constructions & 63A & Noun Phrase Conjunction \\
64A & Nominal and Verbal Conjunction & 65A & Perfective/Imperfective Aspect \\
66A & The Past Tense & 67A & The Future Tense \\
68A & The Perfect & 69A & Position of Tense-Aspect Affixes \\
70A & The Morphological Imperative & 71A & The Prohibitive \\
72A & Imperative-Hortative Systems & 73A & The Optative \\
74A & Situational Possibility & 75A & Epistemic Possibility \\
76A & Overlap between Situational and Epistemic Modal Marking & 77A & Semantic Distinctions of Evidentiality \\
78A & Coding of Evidentiality & 79A & Suppletion According to Tense and Aspect \\
79B & Suppletion in Imperatives and Hortatives & 80A & Verbal Number and Suppletion \\
81A & Order of Subject, Object and Verb & 81B & Languages with two Dominant Orders of Subject, Object, and Verb \\
82A & Order of Subject and Verb & 83A & Order of Object and Verb \\
84A & Order of Object, Oblique, and Verb & 85A & Order of Adposition and Noun Phrase \\
86A & Order of Genitive and Noun & 87A & Order of Adjective and Noun \\
88A & Order of Demonstrative and Noun & 89A & Order of Numeral and Noun \\
90A & Order of Relative Clause and Noun & 90B & Prenominal relative clauses \\
90C & Postnominal relative clauses & 90D & Internally-headed relative clauses \\
90E & Correlative relative clauses & 90F & Adjoined relative clauses \\
90G & Double-headed relative clauses & 91A & Order of Degree Word and Adjective \\
92A & Position of Polar Question Particles & 93A & Position of Interrogative Phrases in Content Questions \\
94A & Order of Adverbial Subordinator and Clause & 95A & Relationship between the Order of Object and Verb and the Order of Adposition and Noun Phrase \\
96A & Relationship between the Order of Object and Verb and the Order of Relative Clause and Noun & 97A & Relationship between the Order of Object and Verb and the Order of Adjective and Noun \\
98A & Alignment of Case Marking of Full Noun Phrases & 99A & Alignment of Case Marking of Pronouns \\
100A & Alignment of Verbal Person Marking & 101A & Expression of Pronominal Subjects \\
102A & Verbal Person Marking & 103A & Third Person Zero of Verbal Person Marking \\
104A & Order of Person Markers on the Verb & 105A & Ditransitive Constructions: The Verb 'Give' \\
106A & Reciprocal Constructions & 107A & Passive Constructions \\
108A & Antipassive Constructions & 108B & Productivity of the Antipassive Construction \\
109A & Applicative Constructions & 109B & Other Roles of Applied Objects \\
110A & Periphrastic Causative Constructions & 111A & Nonperiphrastic Causative Constructions \\
112A & Negative Morphemes & 113A & Symmetric and Asymmetric Standard Negation \\
114A & Subtypes of Asymmetric Standard Negation & 115A & Negative Indefinite Pronouns and Predicate Negation \\
116A & Polar Questions & 117A & Predicative Possession \\
118A & Predicative Adjectives & 119A & Nominal and Locational Predication \\
120A & Zero Copula for Predicate Nominals & 121A & Comparative Constructions \\
122A & Relativization on Subjects & 123A & Relativization on Obliques \\
124A & 'Want' Complement Subjects & 125A & Purpose Clauses \\
126A & 'When' Clauses & 127A & Reason Clauses \\
128A & Utterance Complement Clauses & 129A & Hand and Arm \\
130A & Finger and Hand & 130B & Cultural Categories of Languages with Identity of 'Finger' and 'Hand' \\
131A & Numeral Bases & 132A & Number of Non-Derived Basic Colour Categories \\
133A & Number of Basic Colour Categories & 134A & Green and Blue \\
135A & Red and Yellow & 136A & M-T Pronouns \\
136B & M in First Person Singular & 137A & N-M Pronouns \\
137B & M in Second Person Singular & 138A & Tea \\
139A & Irregular Negatives in Sign Languages & 140A & Question Particles in Sign Languages \\
141A & Writing Systems & 142A & Para-Linguistic Usages of Clicks \\
143A & Order of Negative Morpheme and Verb & 143B & Obligatory Double Negation \\
143C & Optional Double Negation & 143D & Optional Triple Negation \\
143E & Preverbal Negative Morphemes & 143F & Postverbal Negative Morphemes \\
143G & Minor morphological means of signaling negation & 144A & Position of Negative Word With Respect to Subject, Object, and Verb \\
144B & Position of negative words relative to beginning and end of clause and with respect to adjacency to verb & 144C & Languages with different word order in negative clauses \\
144D & The Position of Negative Morphemes in SVO Languages & 144E & Multiple Negative Constructions in SVO Languages \\
144F & Obligatory Double Negation in SVO languages & 144G & Optional Double Negation in SVO languages \\
144H & NegSVO Order & 144I & SNegVO Order \\
144J & SVNegO Order & 144K & SVONeg Order \\
144L & The Position of Negative Morphemes in SOV Languages & 144M & Multiple Negative Constructions in SOV Languages \\
144N & Obligatory Double Negation in SOV languages & 144O & Optional Double Negation in SOV languages \\
144P & NegSOV Order & 144Q & SNegOV Order \\
144R & SONegV Order & 144S & SOVNeg Order \\
144T & The Position of Negative Morphemes in Verb-Initial Languages & 144U & Double negation in verb-initial languages \\
144V & Verb-Initial with Preverbal Negative & 144W & Verb-Initial with Negative that is Immediately Postverbal or between Subject and Object \\
144X & Verb-Initial with Clause-Final Negative & 144Y & The Position of Negative Morphemes in Object-Initial Languages \\
\bottomrule
\end{longtable}
\endgroup
\FloatBarrier

\FloatBarrier
\section{Ablations}\label{app:ablate}
\FloatBarrier
\begin{table}[h]
\centering\small
\caption{Feature-source comparison under identical per-source tuning, from Experiment E1.
The LOLO ordering is preserved on the pure-unseen-language metric, so the Grambank-only advantage is not overfitting.
The LOLO columns use each source's production configuration; the unseen-language column uses the per-source sweep winner.
WALS runs at $n{=}22$; Grambank and Grambank+WALS run at $n{=}24$.}\label{tab:sources}
\begin{tabular}{lcccc}
\toprule
Feature source & \#feat & LOLO $\rho$ & LOLO $R^2$ & unseen-lang $\rho$ \\
\midrule
Grambank (GB)              & 195 & \textbf{0.780} & \textbf{0.607} & \textbf{0.494} \\
WALS                       & 192 & 0.648 & 0.412 & 0.307 \\
Grambank+WALS (operational)& 387 & 0.705 & 0.492 & 0.405 \\
\bottomrule
\end{tabular}
\end{table}

\paragraph{Shared-language source comparison}
Table~\ref{tab:sources} compares Grambank at $n{=}24$ with WALS at $n{=}22$, so its rows rest on different language sets.
Table~\ref{tab:shared22} removes this caveat: all three sources are refit on the 22 languages that the WALS mapping covers, each after its own \texttt{max\_features} mini-sweep, on identical LOLO folds.
The ordering Grambank $>$ Grambank+WALS $>$ WALS is preserved, at $\rho{=}0.716$, $0.673$ and $0.648$.
The Grambank advantage therefore does not depend on the language set.
The production \texttt{max\_features}$=0.5$, tuned at $n{=}24$, drops to $0.576$ on the shared set.
Removing two languages removes 90 directed pairs, from 552 down to 462, so a configuration tuned on the larger set no longer suits the smaller training problem.
This is why each source carries its own sweep on its own language set.
\begin{table}[h]
\centering\small
\caption{Feature-source comparison on the 22 languages shared with WALS (Experiment E1), each source after its own \texttt{max\_features} mini-sweep on identical LOLO folds.
The order of Table~\ref{tab:sources} is preserved on identical languages.}\label{tab:shared22}
\begin{tabular}{lccc}
\toprule
Feature source & best \texttt{max\_features} & LOLO $\rho$ & $R^2$ \\
\midrule
Grambank       & 0.3  & \textbf{0.716} & \textbf{0.511} \\
Grambank+WALS  & sqrt & 0.673 & 0.447 \\
WALS           & log2 & 0.648 & 0.412 \\
\bottomrule
\end{tabular}
\end{table}

\paragraph{WALS-block permutation control}
Tables~\ref{tab:sources} and~\ref{tab:shared22} show the union trailing Grambank under matched tuning.
The permutation control asks whether that drop is generic column dilution at $n{=}24$, or a property of the real WALS data.
The WALS block of the feature matrix is permuted across the 24 languages, so each language receives another language's real WALS profile.
This destroys the feature-to-language association, while every feature's marginal levels and missingness are preserved.
Each permuted design is rerun under the same seven-value \texttt{max\_features} mini-sweep on the production LOLO folds.
Scrambling the WALS block recovers most of the union's drop: the per-design best configuration averages $\rho{=}0.751$ with standard deviation $0.011$ over ten permutations, and all ten score above the real union's $0.705$, against Grambank's $0.780$ (Table~\ref{tab:walsperm}).
The union's drop therefore decomposes into roughly $0.03$ of generic dilution, which any matched block of 192 extra columns causes, and roughly $0.05$ that is specific to the real WALS structure.
The forest fits WALS-to-BTS associations within the training folds that do not transfer to the held-out language.
The WALS features are not noise in isolation, since WALS alone reaches $\rho{=}0.648$, but combined with Grambank at $n{=}24$ they subtract held-out signal rather than adding it.
\begin{table}[h]
\centering\small
\caption{WALS-block permutation control (Experiment E1) under identical LOLO folds and a per-design \texttt{max\_features} mini-sweep.
The scrambled rows permute the WALS block across the 24 languages, preserving marginal levels and missingness; the mean and standard deviation are over ten permutations, every one of which scores above the real union.}\label{tab:walsperm}
\begin{tabular}{lccc}
\toprule
Feature design & \#feat & LOLO $\rho$ & $R^2$ \\
\midrule
Grambank                                       & 195 & 0.780 & 0.607 \\
Grambank+WALS, real block                      & 387 & 0.705 & 0.492 \\
Grambank+WALS, block scrambled (10 perms)      & 387 & $0.751\pm0.011$ & $0.561\pm0.016$ \\
\bottomrule
\end{tabular}
\end{table}

\paragraph{Coverage-floor sweep}
The $0.7$ Grambank coverage threshold defines ATLAS-24, so Table~\ref{tab:floor} refits every tier as the floor is lowered stepwise to zero.
Each step admits exactly one further ATLAS language: Telugu at floor $0.60$, Thai at $0.50$, Indonesian at $0.40$, and Central Kurdish at $0.00$.
Each admitted language is not one training example but $2(n{-}1)$ of them, since every ordered pair with an existing language becomes a directed training pair.
The corpus therefore grows quadratically, from 552 pairs at $n{=}24$ to 756 at $n{=}28$.
Moderately lowering the floor raises the tuned RF from $R^2{=}0.49$ at $n{=}24$ to $0.68$ at $n{=}26$ and $n{=}27$.
The untuned RF rises with it, collapsing the tuning gap from $0.14$ to at most $0.04$.
The added languages relieve the overfitting that separates the untuned from the tuned forest.
Removing the floor entirely reverses the gain.
The near-empty Central Kurdish profile, with Grambank coverage $0.08$, drops the untuned RF to $R^2{=}0.20$ and the tuned RF to $0.48$.
The GLM stays at chance across the entire sweep, so the RF movement tracks the separability of one-hot feature profiles rather than a stronger smooth typological signal.
The pure-unseen-language protocol, whose test pairs share no language with training, shows that the moderate floors also improve generalization.
Unseen-language $\rho$ rises from $0.405$ at floor $0.70$ to $0.508$ at $0.50$ and $0.514$ at $0.40$, nearly doubling $R^2$ from $0.136$ to $0.264$.
Adequately attested extra languages therefore improve prediction of languages the model has never seen, not just the LOLO score.
At floor $0.00$ the same protocol collapses to $\rho{=}{-}0.012$ with $R^2{=}{-}0.20$, below the mean predictor.
The near-empty profile does not merely add noise, it breaks extrapolation entirely.
A one-language change can also move the RF substantially.
As throughout, at $n{=}24$ every language appears in training through other pairs, so the tuned column reads as threshold sensitivity, while the unseen-language column isolates the floor's role: excluding profiles too sparse to extrapolate from.
\begin{table}[h]
\centering\small
\caption{Coverage-floor sweep of the $0.7$ Grambank threshold that defines ATLAS-24.
Every tier is refit at each floor under the production configuration, and each lowering admits the listed language, reaching all 28 ATLAS languages with a Grambank profile at floor $0.00$.
The remaining ten of the 38 ATLAS languages, among them Spanish and German, are absent from Grambank and cannot enter at any floor.
$R^2$ is LOLO true $R^2$; the final two columns are the tuned RF's LOLO $\rho$ and pure-unseen-language $\rho$, whose ATLAS-24 values are the headline $0.705$ and $0.405$.}\label{tab:floor}
\begin{tabular}{lccccccc}
\toprule
Floor & $n$ & language added & glm $R^2$ & full $R^2$ & tuned $R^2$ & tuned $\rho$ & unseen-lang $\rho$ \\
\midrule
$0.70$ & 24 & --              & $0.014$ & $0.350$ & $0.492$ & $0.705$ & $0.405$ \\
$0.60$ & 25 & Telugu          & $0.003$ & $0.507$ & $0.529$ & $0.731$ & $0.411$ \\
$0.50$ & 26 & Thai            & $-0.009$ & $0.658$ & $\mathbf{0.680}$ & $0.824$ & $0.508$ \\
$0.40$ & 27 & Indonesian      & $0.001$ & $0.642$ & $0.676$ & $0.823$ & $\mathbf{0.514}$ \\
$0.00$ & 28 & Central Kurdish & $0.033$ & $0.205$ & $0.476$ & $0.700$ & $-0.012$ \\
\bottomrule
\end{tabular}
\end{table}

\begin{table}[h]
\centering\small
\caption{Matched-capacity and matched-tuning controls from Experiment E1.
A matched-capacity non-typological RF recovers $\rho{\approx}0.58$ at the shared configuration and $0.620$ under its own hyperparameter sweep, well above the single-feature ridges.
Typology's full-capacity increment is $\Delta\rho{=}0.085$ under matched tuning and disappears at a matched five-feature budget, $0.597$ against $0.620$, which locates the advantage in feature breadth.}\label{tab:matched}
\begin{tabular}{lccc}
\toprule
Model (same RF, same folds) & \#feat & LOLO $\rho$ & $R^2$ \\
\midrule
metadata RF (script/family/genus/geo/wiki) & 5   & 0.581 & 0.264 \\
metadata RF, own-sweep tuned               & 5   & 0.620 & 0.375 \\
typology RF (Grambank+WALS)                & 387 & 0.705 & 0.492 \\
typology RF, budget-matched to metadata    & 5   & 0.597 & -- \\
\bottomrule
\end{tabular}
\end{table}

\paragraph{Matched-capacity control details}
The metadata RF in Table~\ref{tab:matched} one-hot encodes, per language, the writing script with 9 levels, the WALS family, the WALS genus, a $k$-means geographic region from lat/lon with 6 clusters, and a Wikipedia article-count bin with 4 quantiles.
Missing values receive an explicit unknown level, so the metadata model is informed of the same missingness the typology model is spared.
This choice is biased in favor of metadata, which is the conservative direction for the comparison.
The budget-matched typology row averages LOLO $\rho$ over 20 random 5-feature subsets drawn from the top informative features under a fixed seed.
Code and per-experiment \texttt{\_meta.json} records, holding the resolved config and git commit, accompany each result CSV.

\paragraph{Matched-tuning control details}
The production metadata control inherits \texttt{max\_features}$=0.5$ from the typology winner, which is conservative for metadata.
We therefore swept the metadata RF over the same curated grid as the typology model.
We screened all 168 configurations under LOLO at 200 trees and seed 42, and refined the top eight over seeds 42 to 46.
The winner by true $R^2$ uses \texttt{max\_features} \texttt{sqrt} with \texttt{max\_depth} 10, reaching LOLO $\rho{=}0.620$ and $R^2{=}0.375$ at seed 42, and $\rho{=}0.602{\pm}0.011$ over the five refinement seeds.
Leave-2-languages-out gives $0.621$ for the winner and $0.611$ for the production configuration, against $0.702$ for typology.
The typology increment is therefore stable across seeds, configurations and holdout protocols, rather than an artifact of unequal tuning.
A paired cluster bootstrap over source languages does not, however, separate the two models at $n{=}24$: $\Delta\rho{=}0.085$ at 95\% CI $[-0.11, 0.41]$, whose width reflects coarse language-level resampling at this scale.
Under fully nested tuning, where both models re-select their configuration inside each training fold, the two converge: typology reaches $\rho{=}0.580$ and metadata $0.571$, a paired difference of $+0.009$ at 95\% CI $[-0.17, 0.24]$.
The same nested procedure scores the typology sweep at $0.580$ against its reported $0.706$, with 21 different configurations winning the 24 inner selections, so per-fold selection is unstable at this scale and tuned numbers read as upper bounds.

\paragraph{Digitization-noise robustness}
The target matrix is read from Figure~C.2 at one decimal, so every BTS cell carries a rounding error of at most $0.05$.
We perturbed each of the 552 targets with independent $U(-0.05, 0.05)$ noise and reran the full tuned-RF LOLO on each noisy target, with the forest seed fixed so that only the labels vary.
Across $B{=}100$ replicates, LOLO $\rho$ averages $0.705$ with standard deviation $0.007$ and worst case $0.688$.
$R^2$ averages $0.492$ with standard deviation $0.011$.
The main reported correlation is therefore insensitive to digitization error at the reading precision.

\begin{figure}[h]
\centering
\includegraphics[width=\linewidth]{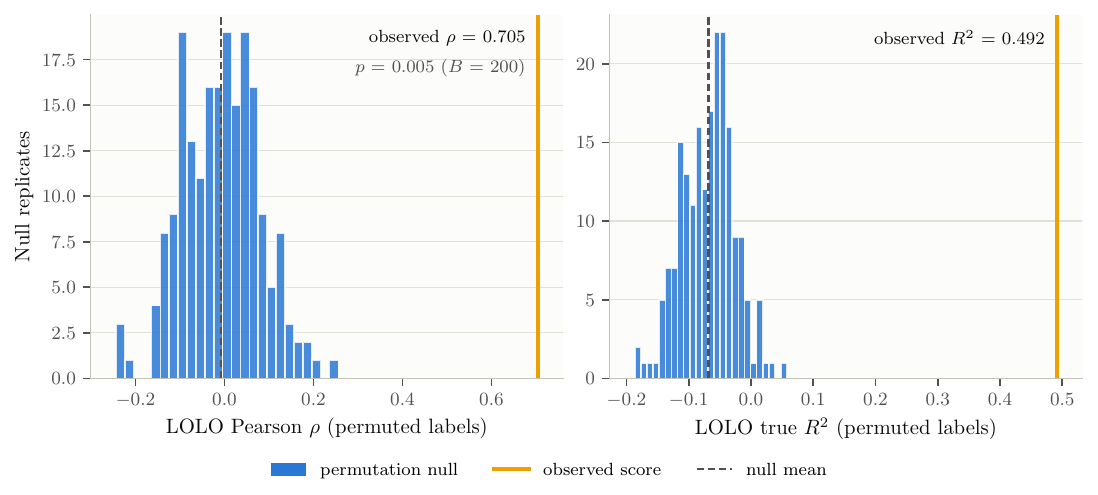}
\caption{Label-permutation null versus observed LOLO scores for the tuned random forest on Grambank+WALS features (ATLAS-24, $B{=}200$ replicates, seed 42).
Left: Pearson $\rho$; right: $R^2$.
Bars show the fold-respecting permutation null, the dashed line its mean, and the solid line the observed score.
The observed $\rho$ exceeds all 200 null replicates ($p{=}0.005$).}\label{fig:permutation_null}
\end{figure}

\begin{table}[h]
\centering\small
\caption{Symmetric distance regressions on the $24{\times}24$ upper triangle, the cross-check within Experiment E3.
Script is a strong marginal correlate but provides no incremental directed power, since the script-only directed model is near chance.}\label{tab:symmetric}
\begin{tabular}{lcccc}
\toprule
Factor & Mantel $r$ & Mantel $p$ & MRM coef & MRM $p$ \\
\midrule
typological   & 0.266 & 0.015 & 0.249 & 0.281 \\
script        & 0.334 & 0.017 & 0.149 & 0.035 \\
genealogical  & 0.226 & 0.051 & 0.098 & 0.693 \\
geographic    & 0.169 & 0.078 & 0.029 & 0.803 \\
\bottomrule
\end{tabular}
\end{table}

\paragraph{Composition-matched grouped holdout (E3)}
Macro-$\rho$ under LOSO and LOFO averages held-out sets that are dominated by cross-script pairs, where every tier is stronger, so the pooled LOLO value of $0.705$ is not the right baseline for them.
Table~\ref{tab:composition} scores pooled LOLO predictions on exactly the pairs each group contributes.
Grouped holdout exceeds this matched baseline on the macro average for both axes, and matches it under pair weighting.
Individual groups move in both directions.
The Arabic script dips to $\rho{=}0.459$ against $0.683$ matched, and Latin to $0.597$ against $0.692$, while the single-language script groups improve markedly: Greek at $0.935$ against $0.572$, and Hebrew at $0.914$ against $0.550$.
For a single-language group the grouped fold is that language's own LOLO fold, so its gain over the matched column reflects the mixed fold provenance of pooled predictions rather than a training difference.
No group approaches the script-only model at $\rho{=}0.138$, so the dips are attenuations, not a collapse.
\begin{table}[h]
\centering\small
\caption{Composition-matched control for the grouped holdouts of Experiment E3.
Grouped $\rho$ retrains with every pair touching the group held out; matched LOLO scores the pooled LOLO predictions on exactly that group's held-out pairs.
Macro rows are unweighted means over groups; weighted rows average by held-out pair count.}\label{tab:composition}
\begin{tabular}{llcc}
\toprule
Axis & Group & grouped $\rho$ & matched LOLO $\rho$ \\
\midrule
LOSO & Arabic     & 0.459 & 0.683 \\
LOSO & Cyrillic   & 0.935 & 0.749 \\
LOSO & Devanagari & 0.912 & 0.622 \\
LOSO & Greek      & 0.935 & 0.572 \\
LOSO & Han        & 0.730 & 0.819 \\
LOSO & Hangul     & 0.793 & 0.643 \\
LOSO & Hebrew     & 0.914 & 0.550 \\
LOSO & Japanese   & 0.708 & 0.676 \\
LOSO & Latin      & 0.597 & 0.692 \\
\midrule
LOSO & macro      & \textbf{0.776} & 0.667 \\
LOSO & weighted   & 0.679 & 0.683 \\
\midrule
LOFO & Afro-Asiatic    & 0.900 & 0.718 \\
LOFO & Altaic          & 0.838 & 0.676 \\
LOFO & Austro-Asiatic  & 0.243 & 0.267 \\
LOFO & Austronesian    & 0.892 & 0.801 \\
LOFO & Indo-European   & 0.693 & 0.723 \\
LOFO & Japanese        & 0.708 & 0.676 \\
LOFO & Korean          & 0.793 & 0.643 \\
LOFO & Niger-Congo     & 0.760 & 0.715 \\
LOFO & Sino-Tibetan    & 0.730 & 0.819 \\
\midrule
LOFO & macro      & \textbf{0.729} & 0.671 \\
LOFO & weighted   & 0.721 & 0.697 \\
\bottomrule
\end{tabular}
\end{table}

\paragraph{Leak-free top-$k$ control (E4)}
The sweep's global feature ranking is scored on the same LOLO folds that are used for evaluation, so its $k{<}387$ entries are optimistically biased.
In the control, fold $i$ selects its own top-$k$ by an inner LOLO importance computed entirely within fold $i$'s training pairs, so the held-out language touches neither the fit nor the ranking.
The $k{=}200$ value falls only from $0.742$ to $0.734$.
Across three seeds, the full model scores $\rho=0.708\pm0.004$, while leak-free $k{=}200$ scores $0.730\pm0.004$, so the peak survives honest within-fold selection.
Per-fold selection also beats the global ranking across the mid-range, for example $0.690$ versus $0.627$ at $k{=}100$ and $0.692$ versus $0.573$ at $k{=}10$, yet it never significantly beats the full 387-feature model; only the single best feature ($k{=}1$, $0.467$) drops, consistent with the instability of the top-25 ranking.

\paragraph{Inferential statistics.}
Table~\ref{tab:infer} collects the permutation and bootstrap statistics behind the main results of E1, E3, E4, and E5.
\begin{table}[h]
\centering\small
\caption{Inferential statistics, with each row labeled by the experiment and research question it addresses.
The permutation null uses $B{=}200$, and the cluster and paired bootstraps use $B{=}2000$.
The final row's interval is the permuted-label null's central 95\% range, not a bootstrap interval.}\label{tab:infer}
\begin{tabular}{lllcc}
\toprule
Exp. & RQ & Quantity & Estimate & 95\% CI \\
\midrule
E1 & RQ1 & LOLO $\rho$; permutation $p{=}0.005$            & $0.705$ & $[0.51, 0.82]$ \\
E1 & RQ1 & LOLO $\rho$; language-level bootstrap 95\% CI   & $0.705$ & $[0.50, 0.85]$ \\
E1 & RQ1 & pure-unseen-language $\rho$ (Grambank+WALS)     & $0.405$ & -- \\
E3 & RQ1 & LOSO / LOFO macro-$\rho$                        & $0.776$ / $0.729$ & $[0.67,0.88]$ / $[0.59,0.83]$ \\
E4 & RQ1 & top-25 importance Jaccard (random ${\approx}0.033$) & $0.073$ & $[0.069, 0.078]$ \\
E5 & RQ2 & residual gap, Marathi$-$English (bootstrap over targets); $p{<}0.001$    & $+0.236$ & $[0.19, 0.28]$ \\
E5 & RQ2 & residual margin; null $p{=}0.005$              & $0.246$ & $[0.065,0.173]$ \\
\bottomrule
\end{tabular}
\end{table}

\paragraph{Residual-ranking validity controls}
Two controls ask whether the E5 residual flip is meaningful, or whether the operationalization produces decisive-looking rankings from nothing.
The first removes the in-sample optimism of the bias layer, which explains $R^2{=}0.12$ in-sample but $-0.006$ out-of-fold under the LOLO folds.
Refitting $\hat B$ out-of-fold and recomputing the ranking, both as the raw mean residual and through the production rf$_T$, leaves the flip intact: Marathi remains first at $0.36$, English remains at rank 9 or 10, and both rankings correlate with the production ranking at Spearman $0.99$.
The flip is therefore carried by the nuisance association itself, which replicates across folds at out-of-fold Pearson $0.16$, rather than by the layer's in-sample optimism.
The second control runs the entire operationalization on permuted BTS labels, refitting the bias ridge, the residual, the random forest, and the ranking per replicate.
This quantifies the rankings the pipeline produces from meaningless labels.
The null margin of the top source over the median source averages $0.116$ with interval $[0.065, 0.173]$ over $B{=}200$ replicates, and no replicate reached the observed $0.246$, so the ranking's sharpness exceeds the null at $p{=}0.005$.
The top-source gap over English specifically is weaker evidence: 11 of 200 null replicates exceed the observed rf$_T$ gap of $0.237$ at $p{=}0.055$, because a random draw can seat English at the bottom of a null ranking.
The residual ranking is thus sharper than what the operationalization produces on meaningless labels and stable under honest bias estimation.
The identity of the top source is however weakly determined: the 200 permuted reruns place 24 different languages at rank one.
E5 therefore reports the flip as a diagnostic rather than a recommendation.

\paragraph{Additional distance baselines}
We compare the typology random forest against five further distance baselines drawn from the transfer and tokenization literature \cite{littell-etal-2017-uriel,NEURIPS2023_74bb24dc}.
The lang2vec baseline scores each pair by the cosine of the concatenated imputed featural vectors of \cite{littell-etal-2017-uriel}, covering syntax, phonology and inventory.
The URIEL baseline feeds the six canonical distances of the same knowledge base, namely genetic, geographic, syntactic, phonological, inventory and their featural aggregate, as a six-column design to the same within-fold ridge as the metadata baselines.
The geographic distance is the great-circle distance on Glottolog coordinates, because the lang2vec geographic feature is a vector of distances to fixed reference points rather than per-language coordinates.
Three corpus-lexical baselines measure the Jaccard overlap between per-language artifacts of Wikipedia lead-paragraph corpora of 2{,}000 random articles per edition.
These are byte-level BPE vocabularies of 8{,}000 entries trained per language following \cite{sennrich-etal-2016-neural}, pooled character one-to-three-gram sets over the 5{,}000 most frequent word types in the tradition of \cite{cavnar1994n}, and the 10{,}000 most frequent word types.
The last two follow the tokenizer analysis of \cite{NEURIPS2023_74bb24dc}.
Table~\ref{tab:extrabaselines} scores all five under the protocol of the main text; every tier covers all 24 languages.
All five sit at or below a true $R^2$ of $0.044$.
The character $n$-gram and subword-tokenizer overlaps lead at $\rho$ of $0.225$ and $0.222$, followed by the lang2vec cosine at $0.148$, the word-vocabulary overlap at $0.142$, and the URIEL ridge at $0.095$, which is at chance.
The typology random forest explains $0.49$ of held-out variance, eleven times the best distance baseline.
The script-adjacent lexical overlaps outscore the typological-distance baselines, which matches the symmetric analysis, where script is the strongest marginal correlate of transfer.
None of the five approaches the full typological feature set.
\begin{table}[h]
\centering\small
\caption{Additional distance baselines under the leave-one-language-out protocol of Table~\ref{tab:tiers}, on ATLAS-24.
Every tier covers all 24 languages.
The reference row repeats the tuned typology random forest from the main text.}\label{tab:extrabaselines}
\begin{tabular}{lcc}
\toprule
Baseline & LOLO $\rho$ & true $R^2$ \\
\midrule
character $n$-gram overlap      & 0.225 & 0.044 \\
subword tokenizer overlap       & 0.222 & 0.044 \\
lang2vec featural cosine        & 0.148 & 0.012 \\
word vocabulary overlap         & 0.142 & 0.010 \\
URIEL six-distance ridge        & 0.095 & $-0.019$ \\
\midrule
tuned RF (typology, reference) & 0.705 & 0.492 \\
\bottomrule
\end{tabular}
\end{table}

\FloatBarrier
\section{Transfer matrices}\label{app:transfer}
\FloatBarrier
This section visualizes the transfer matrices behind the ATLAS-24 evaluation.
Every panel is a $38\times38$ heatmap on the axis order of the digitized ATLAS Figure~C.2.
The cell in row $i$, column $j$ is the bilingual transfer score (BTS) for transferring from the column language, the source, to the row language, the target.
The diagonal, self-transfer, is undefined, and the 14 ATLAS languages without Grambank coverage of at least $0.70$ are masked in grey.
Axis labels are colored by language family, at the granularity of the original figure's legend.
All BTS panels, ground truth and four model tiers, share one diverging color scale spanning the ground-truth range [$-1.60$, $0.50$] BTS.
Its neutral point is the midpoint of that range, not zero.
Model panels show leave-one-language-out predictions.
Each cell is produced by a training fold that excludes all pairs involving one of the cell's two languages, so no cell is an in-sample fit.
The difference panels share one symmetric scale of $\pm1.35$ BTS.
Red means the model underpredicts the ground truth, blue means it overpredicts.

\begin{figure}[h]
\centering
\includegraphics[width=\linewidth]{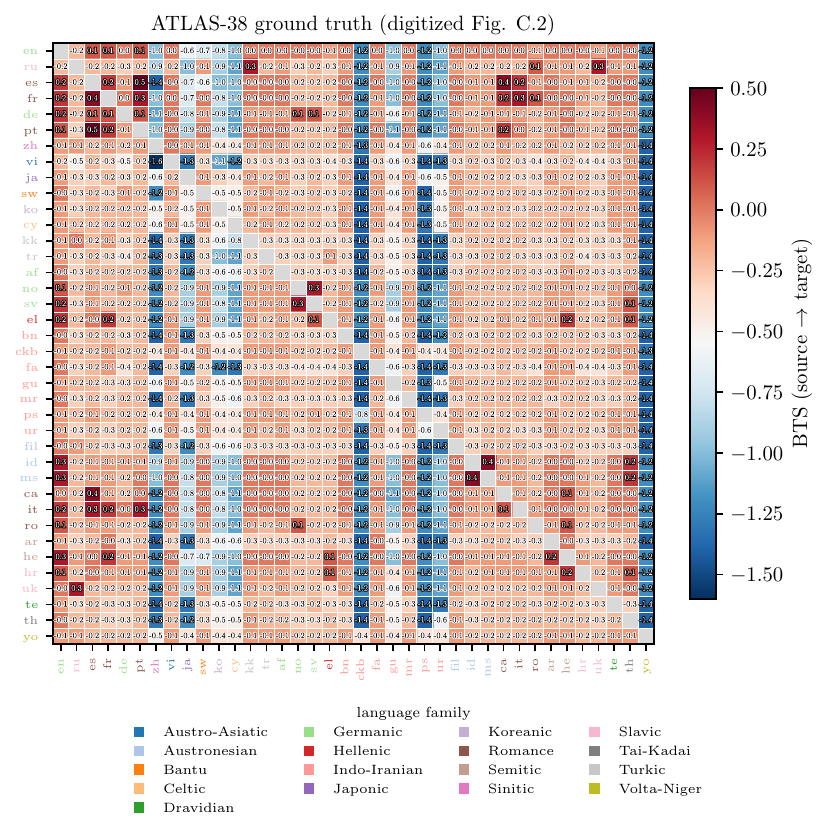}
\caption{Digitized ATLAS-38 ground-truth transfer matrix (Figure~C.2 of the
ATLAS paper). This axis order, family coloring, and BTS color scale are shared
by all model and difference panels in this section.}\label{fig:tm_gt}
\end{figure}

\begin{figure}[h]
\centering
\includegraphics[width=\linewidth]{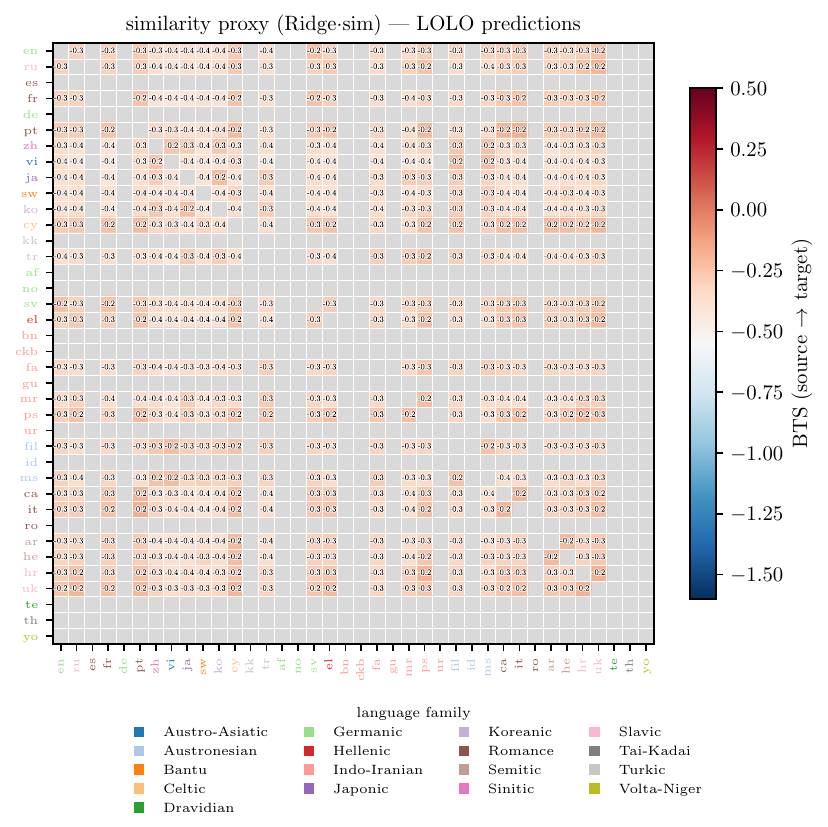}
\caption{LOLO predictions of the similarity tier (ridge on typological
similarity alone). Held-out Pearson $\rho=0.037$, MAE
$=0.277$~BTS.}\label{fig:tm_similarity}
\end{figure}

\begin{figure}[h]
\centering
\includegraphics[width=\linewidth]{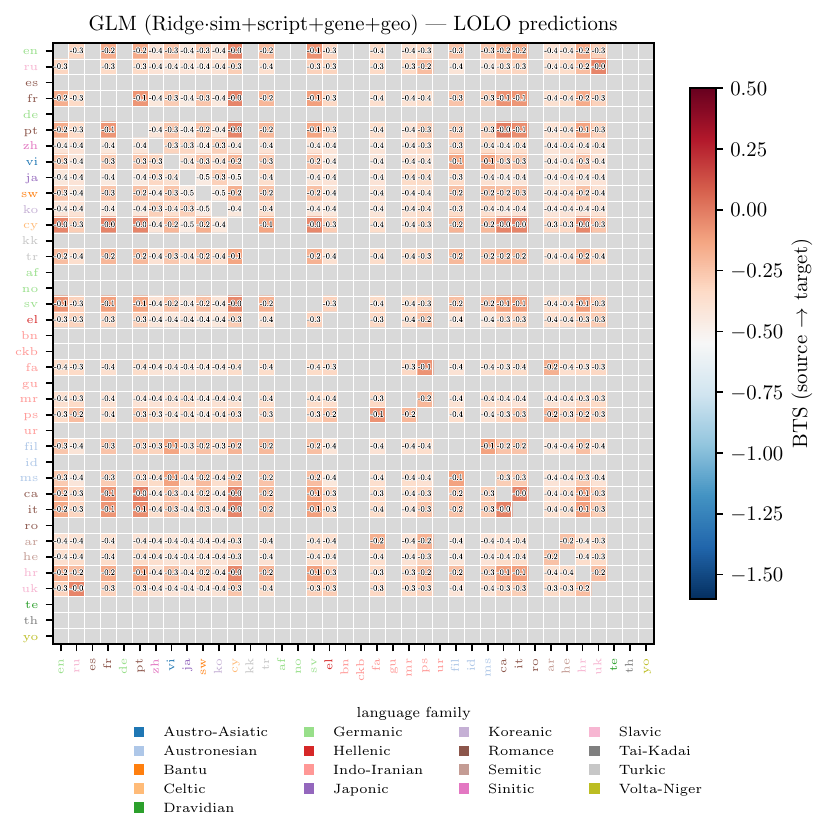}
\caption{LOLO predictions of the GLM tier (ridge on similarity, script,
genealogical, and geographic distances). Held-out Pearson $\rho=0.162$,
MAE $=0.277$~BTS.}\label{fig:tm_glm}
\end{figure}

\begin{figure}[h]
\centering
\includegraphics[width=\linewidth]{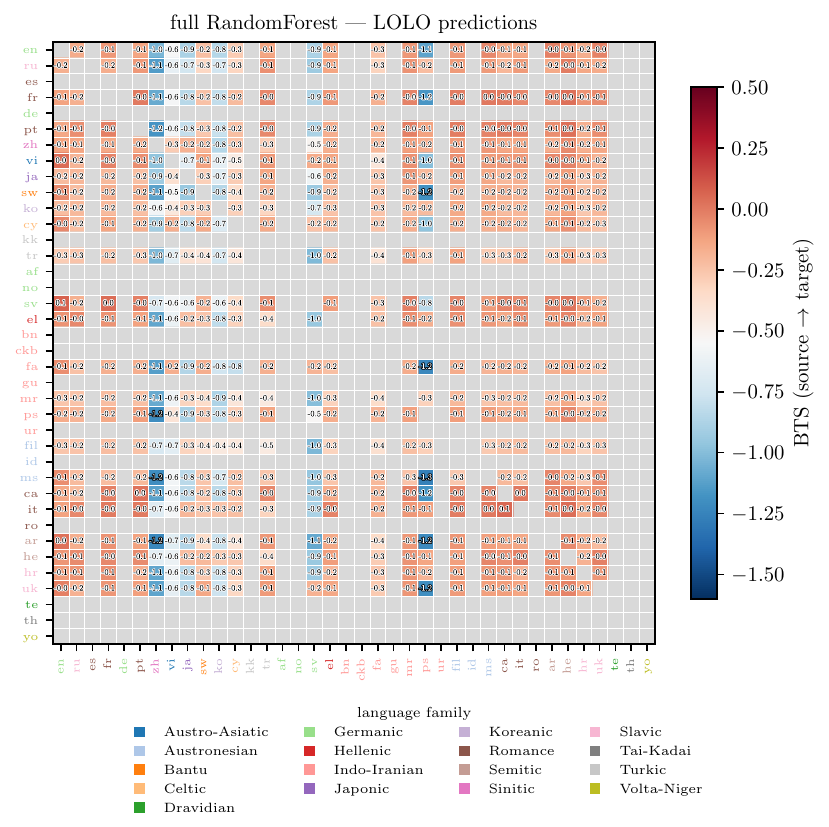}
\caption{LOLO predictions of the full random forest (all typological features,
default hyperparameters). Held-out Pearson $\rho=0.615$, MAE
$=0.203$~BTS.}\label{fig:tm_rf_full}
\end{figure}

\begin{figure}[h]
\centering
\includegraphics[width=\linewidth]{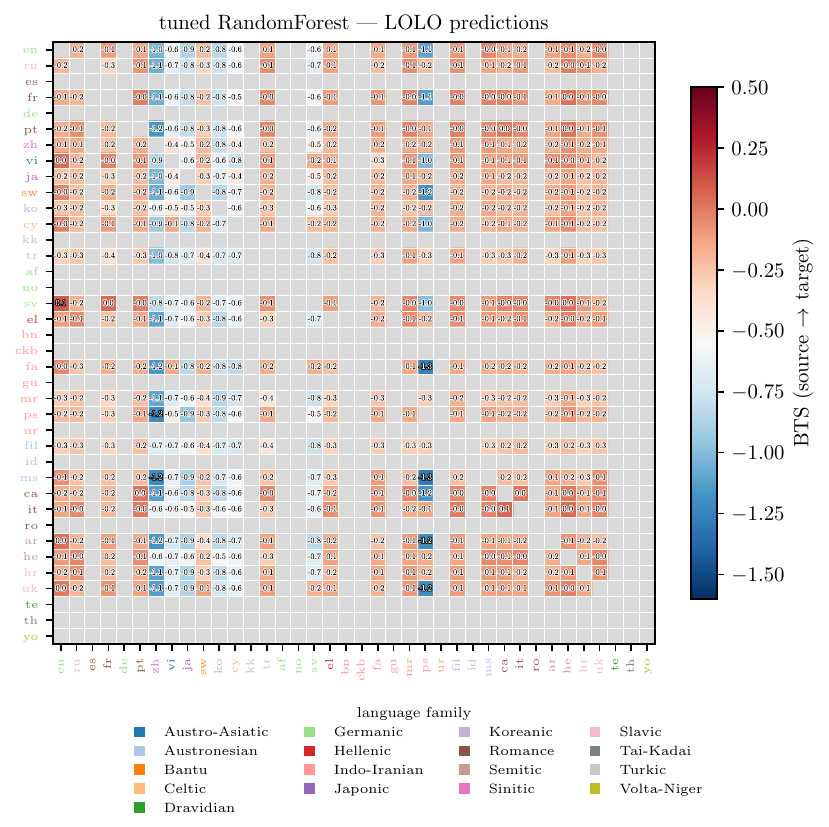}
\caption{LOLO predictions of the tuned random forest (the production model of
the main experiments; max\_features $= 0.5$). Held-out Pearson
$\rho=0.705$, MAE $=0.185$~BTS -- this is the
$\rho=0.705$ of the main ATLAS-24 result.}\label{fig:tm_rf_tuned}
\end{figure}

\begin{figure}[h]
\centering
\includegraphics[width=\linewidth]{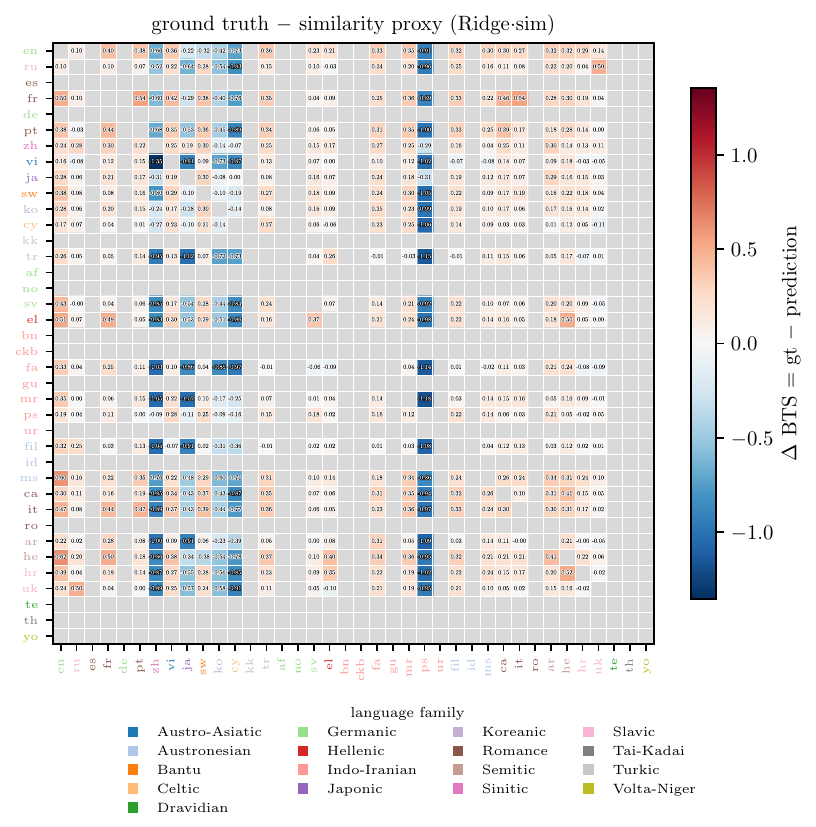}
\caption{Ground truth minus the LOLO predictions of the similarity tier, on the
shared symmetric difference scale. MAE $0.277$~BTS.}\label{fig:tm_diff_similarity}
\end{figure}

\begin{figure}[h]
\centering
\includegraphics[width=\linewidth]{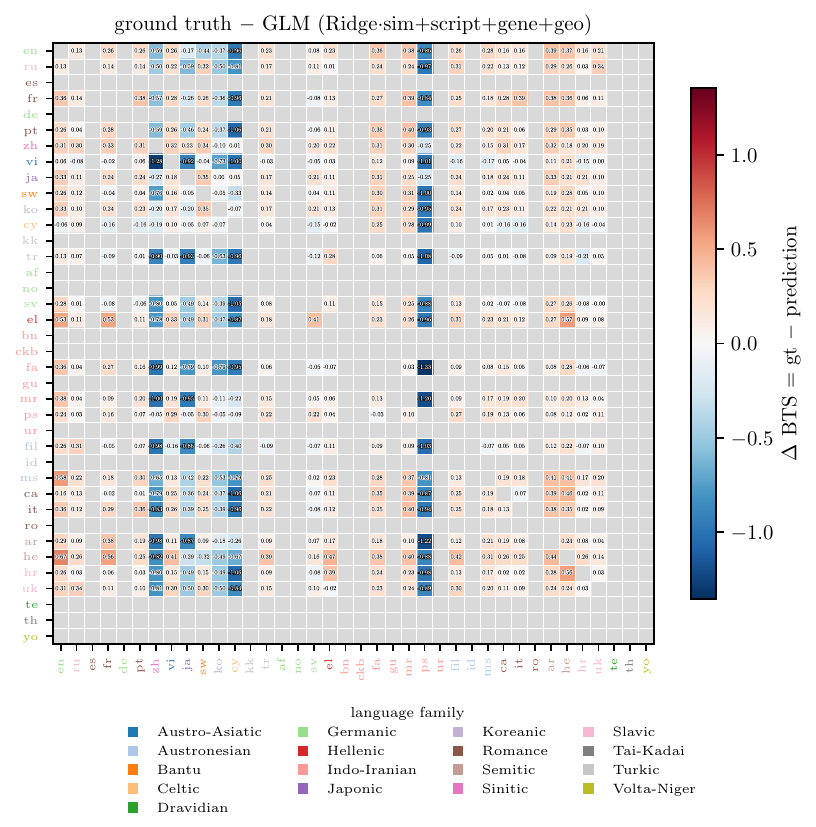}
\caption{Ground truth minus the LOLO predictions of the GLM tier. MAE
$0.277$~BTS.}\label{fig:tm_diff_glm}
\end{figure}

\begin{figure}[h]
\centering
\includegraphics[width=\linewidth]{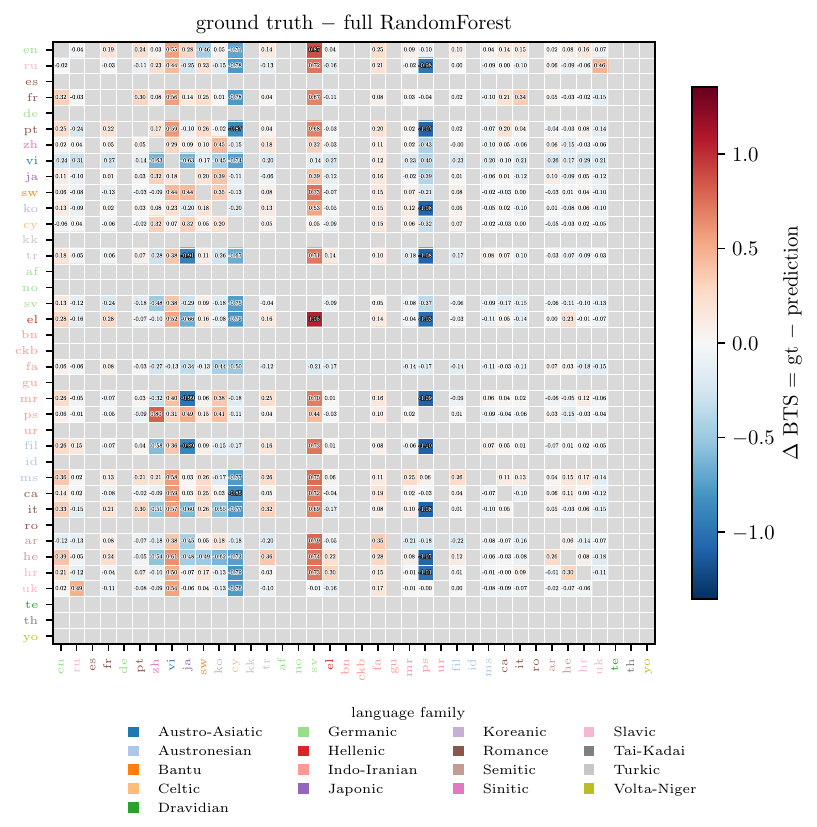}
\caption{Ground truth minus the LOLO predictions of the full random forest.
MAE $0.203$~BTS.}\label{fig:tm_diff_rf_full}
\end{figure}

\begin{figure}[h]
\centering
\includegraphics[width=\linewidth]{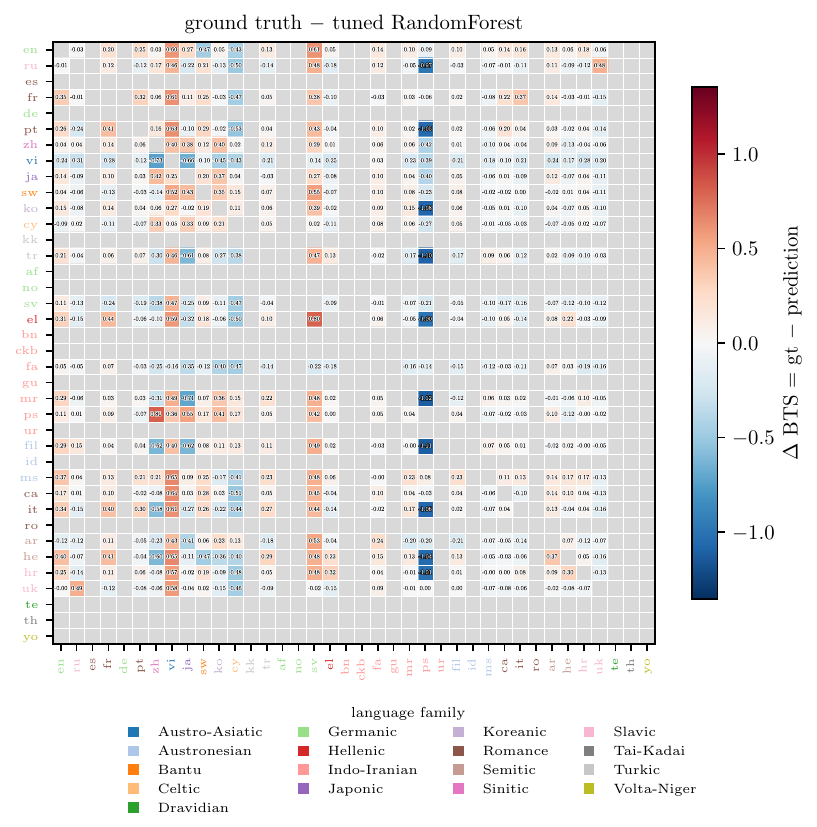}
\caption{Ground truth minus the LOLO predictions of the tuned random forest.
MAE $0.185$~BTS. Note the shrinkage of the error relative to the
weaker tiers: held-out error concentrates near the scale's neutral
midpoint.}\label{fig:tm_diff_rf_tuned}
\end{figure}
\FloatBarrier

\FloatBarrier
\section{Additional results}\label{app:adds}
\FloatBarrier
\begin{table}[h]
\centering\small
\caption{Terms of the bias layer $\hat B$, the ridge fit of BTS on five non-typological nuisance descriptors over the 552 ATLAS-24 directed pairs.
Weights are on the standardized feature scale, and the mean absolute contribution is each feature's contribution to $\hat B$ averaged over pairs.
The layer explains $R^2{=}0.12$ of BTS in-sample at correlation $0.35$, but approximately zero under leave-one-language-out cross-validation.
Its weights are therefore descriptive of ATLAS-24, not a stable law.
Its mean equals the BTS mean of $-0.31$, so the residual $T=\mathrm{BTS}-\hat B$ is mean-zero.
The source-resource and script terms dominate, while the target-resource term is negligible, so the bias is driven by source-side resource and script sharing.}\label{tab:bias_terms}
\begin{tabular}{lcc}
\toprule
Nuisance descriptor & std.\ weight & mean $|$contribution$|$ (BTS) \\
\midrule
source resource ($\log$)      & $+0.087$ & $0.070$ \\
script difference             & $-0.077$ & $0.071$ \\
genealogical distance         & $-0.034$ & $0.027$ \\
geographic distance           & $-0.018$ & $0.016$ \\
target resource ($\log$)      & $+0.002$ & $0.002$ \\
\bottomrule
\end{tabular}
\end{table}

\begin{table}[h]
\centering\small
\caption{Exhaustive feature-subset ablation of the ATLAS-24 bias layer $B$, grouped by subset size $k$.
Each row refits $B$ on that subset of the five nuisance descriptors and propagates to the resource-adjusted residual $T=\mathrm{BTS}-B$.
Abbreviations: scr=script difference, gene=genealogical distance, geo=geographic distance, res$_s$/res$_t$=$\log1p$ Wikipedia resource of source/target.
$R^2(B,y)$ is the bias-explained variance of BTS, and $\|\Delta T\|$ is the relative shift of $T$ from the full-five residual.
LOLO $\rho$ is the leave-one-language-out Pearson of the resource-adjusted BTS predictor, an RF on $T$ with the bias $B$ added back.
The production biased baseline is $\rho=0.705$, and the no-bias-layer row refits within this pipeline, landing at $0.701$.
Rank $\rho$ is the Spearman correlation of the per-source pretraining ranking against the full-model ranking and against empirical BTS ground truth.
The top source is English under every configuration; $\neq$ marks a flip, and none is observed.}\label{tab:bias_ablation}
\resizebox{\linewidth}{!}{%
\begin{tabular}{l r r r r r l}
\toprule
Bias features & $R^2(B,y)$ & $\|\Delta T\|$ (\%) & LOLO $\rho$ & rank $\rho$ (full) & rank $\rho$ (truth) & top \\
\midrule
\multicolumn{7}{l}{\textit{$k=1$ (1 feature)}} \\
geo & 0.013 & 34.8 & 0.719 & 0.989 & 0.984 & English \\
res$_s$ & 0.060 & 26.0 & 0.718 & 0.990 & 0.987 & English \\
scr & 0.051 & 28.0 & 0.717 & 0.979 & 0.984 & English \\
gene & 0.023 & 33.1 & 0.712 & 0.992 & 0.988 & English \\
res$_t$ & 0.000 & 36.8 & 0.705 & 0.989 & 0.990 & English \\
\midrule
\multicolumn{7}{l}{\textit{$k=2$ (2 features)}} \\
scr, res$_s$ & 0.106 & 12.7 & 0.748 & 0.989 & 0.987 & English \\
scr, gene & 0.066 & 24.8 & 0.742 & 0.978 & 0.982 & English \\
scr, geo & 0.063 & 25.3 & 0.734 & 0.978 & 0.974 & English \\
gene, res$_s$ & 0.079 & 21.4 & 0.729 & 0.983 & 0.977 & English \\
gene, res$_t$ & 0.023 & 33.1 & 0.723 & 0.993 & 0.992 & English \\
scr, res$_t$ & 0.051 & 28.0 & 0.722 & 0.983 & 0.984 & English \\
gene, geo & 0.025 & 32.7 & 0.720 & 0.993 & 0.990 & English \\
geo, res$_t$ & 0.013 & 34.8 & 0.717 & 0.994 & 0.988 & English \\
geo, res$_s$ & 0.068 & 24.2 & 0.713 & 0.987 & 0.982 & English \\
res$_s$, res$_t$ & 0.061 & 25.9 & 0.704 & 0.995 & 0.990 & English \\
\midrule
\multicolumn{7}{l}{\textit{$k=3$ (3 features)}} \\
scr, gene, res$_s$ & 0.118 & 4.5 & 0.750 & 1.000 & 0.995 & English \\
scr, res$_s$, res$_t$ & 0.106 & 12.6 & 0.746 & 0.995 & 0.993 & English \\
scr, geo, res$_t$ & 0.063 & 25.3 & 0.740 & 0.987 & 0.979 & English \\
scr, geo, res$_s$ & 0.114 & 8.1 & 0.738 & 0.995 & 0.993 & English \\
scr, gene, res$_t$ & 0.066 & 24.8 & 0.734 & 0.989 & 0.989 & English \\
scr, gene, geo & 0.069 & 23.9 & 0.733 & 0.990 & 0.984 & English \\
gene, geo, res$_s$ & 0.080 & 21.3 & 0.729 & 0.984 & 0.978 & English \\
gene, res$_s$, res$_t$ & 0.080 & 21.4 & 0.727 & 0.985 & 0.982 & English \\
geo, res$_s$, res$_t$ & 0.069 & 24.1 & 0.719 & 0.990 & 0.984 & English \\
gene, geo, res$_t$ & 0.025 & 32.7 & 0.714 & 0.996 & 0.995 & English \\
\midrule
\multicolumn{7}{l}{\textit{$k=4$ (4 features)}} \\
scr, gene, res$_s$, res$_t$ & 0.118 & 4.4 & 0.760 & 0.999 & 0.994 & English \\
scr, geo, res$_s$, res$_t$ & 0.114 & 8.1 & 0.757 & 0.998 & 0.994 & English \\
scr, gene, geo, res$_s$ & 0.120 & 0.5 & 0.756 & 0.997 & 0.995 & English \\
scr, gene, geo, res$_t$ & 0.069 & 23.9 & 0.733 & 0.987 & 0.986 & English \\
gene, geo, res$_s$, res$_t$ & 0.080 & 21.2 & 0.723 & 0.991 & 0.985 & English \\
\midrule
\multicolumn{7}{l}{\textit{$k=5$ (5 features)}} \\
\textbf{scr, gene, geo, res$_s$, res$_t$} & 0.120 & 0.0 & 0.747 & 1.000 & 0.995 & English \\
\midrule
\textit{none (no bias layer)} & 0.000 & 36.8 & 0.701 & 0.994 & 0.993 & English \\
\bottomrule
\end{tabular}}
\end{table}

\begin{table}[h]
\centering\small
\caption{Full 24-language ranking of the ATLAS-24 candidate set under each operationalization of Experiment E5.
The candidate set is used as both sources and targets.
The columns are the empirical raw-BTS mean, the resource-adjusted residual $T=\mathrm{BTS}-\hat B$, the mean typological similarity to the targets, and the OOD-generalizer mean BTS to cross-family targets.
Scores are on each column's own scale and are not comparable across columns; only the within-column order is meaningful.
The on-scale reweighted random forest of Table~\ref{tab:rq4} is omitted here because its full 24-language ranking tracks the empirical one at Spearman $0.974$, so its top sources are summarized in the main text only.
The bold top row shows that the four methods disagree on the best source, which is the RQ2 result.}\label{tab:e5_fullrank}
\resizebox{\linewidth}{!}{%
\begin{tabular}{c l r l r l r l r}
\toprule
Rank & \multicolumn{2}{c}{empirical} & \multicolumn{2}{c}{resource-adj.} & \multicolumn{2}{c}{centrality} & \multicolumn{2}{c}{OOD-generalizer} \\
\cmidrule(lr){2-3}\cmidrule(lr){4-5}\cmidrule(lr){6-7}\cmidrule(lr){8-9}
 & source & score & source & score & source & score & source & score \\
\midrule
1 & \textbf{English} & \textbf{0.022} & \textbf{Marathi} & \textbf{0.361} & \textbf{Ukrainian} & \textbf{0.717} & \textbf{English} & \textbf{-0.011} \\
2 & Modern Hebrew & -0.074 & Filipino & 0.355 & Portuguese & 0.700 & Modern Hebrew & -0.065 \\
3 & French & -0.109 & Modern Hebrew & 0.328 & Serbian-Croatian-Bosnian & 0.679 & Filipino & -0.114 \\
4 & Filipino & -0.113 & Swahili & 0.235 & Catalan & 0.678 & Vietnamese & -0.114 \\
5 & Standard Arabic & -0.113 & Standard Arabic & 0.198 & Italian & 0.677 & Standard Arabic & -0.120 \\
6 & Portuguese & -0.113 & Western Farsi & 0.187 & Southern Pashto & 0.673 & Western Farsi & -0.122 \\
7 & Marathi & -0.113 & Modern Greek & 0.180 & Swedish & 0.662 & Turkish & -0.124 \\
8 & Western Farsi & -0.113 & Turkish & 0.152 & English & 0.661 & Marathi & -0.144 \\
9 & Vietnamese & -0.122 & Standard Malay & 0.152 & Modern Greek & 0.659 & French & -0.144 \\
10 & Catalan & -0.122 & Vietnamese & 0.136 & Welsh & 0.659 & Portuguese & -0.156 \\
11 & Turkish & -0.130 & English & 0.125 & French & 0.654 & Catalan & -0.167 \\
12 & Italian & -0.170 & Catalan & 0.115 & Filipino & 0.654 & Standard Malay & -0.171 \\
13 & Standard Malay & -0.174 & Serbian-Croatian-Bosnian & 0.115 & Standard Malay & 0.653 & Swahili & -0.181 \\
14 & Swahili & -0.183 & Portuguese & 0.102 & Russian & 0.647 & Modern Greek & -0.211 \\
15 & Serbian-Croatian-Bosnian & -0.200 & French & 0.057 & Modern Hebrew & 0.647 & Italian & -0.233 \\
16 & Swedish & -0.204 & Ukrainian & 0.051 & Standard Arabic & 0.642 & Swedish & -0.233 \\
17 & Modern Greek & -0.217 & Russian & 0.041 & Western Farsi & 0.623 & Serbian-Croatian-Bosnian & -0.233 \\
18 & Russian & -0.217 & Italian & 0.022 & Marathi & 0.613 & Russian & -0.267 \\
19 & Ukrainian & -0.230 & Swedish & -0.036 & Turkish & 0.611 & Ukrainian & -0.278 \\
20 & Korean & -0.752 & Korean & -0.384 & Mandarin Chinese & 0.604 & Welsh & -0.744 \\
21 & Japanese & -0.848 & Japanese & -0.505 & Korean & 0.590 & Korean & -0.776 \\
22 & Welsh & -0.883 & Welsh & -0.589 & Japanese & 0.579 & Japanese & -0.852 \\
23 & Mandarin Chinese & -1.087 & Southern Pashto & -0.650 & Swahili & 0.575 & Mandarin Chinese & -1.110 \\
24 & Southern Pashto & -1.217 & Mandarin Chinese & -0.736 & Vietnamese & 0.573 & Southern Pashto & -1.209 \\
\bottomrule
\end{tabular}}
\end{table}

\begin{table}[h]
\centering\small
\caption{Full 24-language ranking of the ATLAS-24 candidate set by positive-transfer coverage: the number of target languages with a positive transfer score.
The columns are the empirical raw BTS out of 23 targets, the bias-removed residual $T=\mathrm{BTS}-\hat B$ out of 23 targets, and the OOD-generalizer raw BTS to cross-family targets, out of that source's cross-family target count.
Count ties are broken by the corresponding mean score in Table~\ref{tab:e5_fullrank}.
Counts are not comparable across columns; only the within-column order is meaningful.}
\label{tab:e5_posrank}
\resizebox{\linewidth}{!}{%
\begin{tabular}{c l r l r l r}
\toprule
Rank & \multicolumn{2}{c}{empirical} & \multicolumn{2}{c}{bias-removed} & \multicolumn{2}{c}{OOD-generalizer} \\
\cmidrule(lr){2-3}\cmidrule(lr){4-5}\cmidrule(lr){6-7}
 & source & count & source & count & source & count \\
\midrule
1 & \textbf{English} & \textbf{8/23} & \textbf{Marathi} & \textbf{23/23} & \textbf{Modern Hebrew} & \textbf{3/20} \\
2 & French & 5/23 & Filipino & 23/23 & English & 2/9 \\
3 & Modern Hebrew & 3/23 & Modern Hebrew & 23/23 & French & 1/9 \\
4 & Portuguese & 3/23 & Standard Arabic & 23/23 & Modern Greek & 1/9 \\
5 & Catalan & 3/23 & Western Farsi & 23/23 & Filipino & 0/22 \\
6 & Modern Greek & 2/23 & Modern Greek & 23/23 & Vietnamese & 0/21 \\
7 & Standard Arabic & 1/23 & Standard Malay & 22/23 & Standard Arabic & 0/20 \\
8 & Italian & 1/23 & Swahili & 21/23 & Western Farsi & 0/9 \\
9 & Swedish & 1/23 & Turkish & 21/23 & Turkish & 0/21 \\
10 & Russian & 1/23 & Vietnamese & 21/23 & Marathi & 0/9 \\
11 & Ukrainian & 1/23 & English & 20/23 & Portuguese & 0/9 \\
12 & Filipino & 0/23 & Catalan & 20/23 & Catalan & 0/9 \\
13 & Marathi & 0/23 & Serbian-Croatian-Bosnian & 20/23 & Standard Malay & 0/21 \\
14 & Western Farsi & 0/23 & Portuguese & 19/23 & Swahili & 0/21 \\
15 & Vietnamese & 0/23 & Ukrainian & 17/23 & Italian & 0/9 \\
16 & Turkish & 0/23 & Russian & 16/23 & Serbian-Croatian-Bosnian & 0/9 \\
17 & Standard Malay & 0/23 & French & 15/23 & Swedish & 0/9 \\
18 & Swahili & 0/23 & Italian & 14/23 & Russian & 0/9 \\
19 & Serbian-Croatian-Bosnian & 0/23 & Swedish & 8/23 & Ukrainian & 0/9 \\
20 & Korean & 0/23 & Korean & 0/23 & Welsh & 0/9 \\
21 & Japanese & 0/23 & Japanese & 0/23 & Korean & 0/21 \\
22 & Welsh & 0/23 & Welsh & 0/23 & Japanese & 0/21 \\
23 & Mandarin Chinese & 0/23 & Southern Pashto & 0/23 & Mandarin Chinese & 0/21 \\
24 & Southern Pashto & 0/23 & Mandarin Chinese & 0/23 & Southern Pashto & 0/22 \\
\bottomrule
\end{tabular}}
\end{table}

\begin{table}[h]
\centering\small
\caption{Full 24-language ranking of the ATLAS-24 candidate set by best-source coverage: the number of target languages for which the source attains the highest transfer score.
The columns are the empirical raw BTS, the bias-removed residual $T=\mathrm{BTS}-\hat B$, and the raw BTS restricted to cross-family sources, the best outside-family option per target.
A tie at a target's maximum counts for every co-best source, so column counts sum to more than 24: empirically, 11 of 24 targets have tied maxima, up to 9-way.
Count ties between sources are broken by the corresponding mean score in Table~\ref{tab:e5_fullrank}.
Only the within-column order is meaningful.}
\label{tab:e5_argmax}
\resizebox{\linewidth}{!}{%
\begin{tabular}{c l r l r l r}
\toprule
Rank & \multicolumn{2}{c}{empirical} & \multicolumn{2}{c}{bias-removed} & \multicolumn{2}{c}{OOD-generalizer} \\
\cmidrule(lr){2-3}\cmidrule(lr){4-5}\cmidrule(lr){6-7}
 & source & count & source & count & source & count \\
\midrule
1 & \textbf{English} & \textbf{14/24} & \textbf{Marathi} & \textbf{10/24} & \textbf{Modern Hebrew} & \textbf{14/24} \\
2 & Modern Hebrew & 6/24 & Filipino & 8/24 & Swahili & 10/24 \\
3 & Western Farsi & 5/24 & Modern Hebrew & 6/24 & English & 9/24 \\
4 & French & 4/24 & Swahili & 0/24 & Vietnamese & 9/24 \\
5 & Filipino & 4/24 & Standard Arabic & 0/24 & Standard Arabic & 9/24 \\
6 & Marathi & 4/24 & Western Farsi & 0/24 & Turkish & 8/24 \\
7 & Portuguese & 4/24 & Modern Greek & 0/24 & Western Farsi & 4/24 \\
8 & Swahili & 4/24 & Turkish & 0/24 & Marathi & 3/24 \\
9 & Vietnamese & 3/24 & Standard Malay & 0/24 & Standard Malay & 2/24 \\
10 & Standard Arabic & 2/24 & Vietnamese & 0/24 & French & 1/24 \\
11 & Catalan & 2/24 & English & 0/24 & Portuguese & 1/24 \\
12 & Turkish & 2/24 & Catalan & 0/24 & Catalan & 1/24 \\
13 & Italian & 1/24 & Serbian-Croatian-Bosnian & 0/24 & Modern Greek & 1/24 \\
14 & Swedish & 1/24 & Portuguese & 0/24 & Swedish & 1/24 \\
15 & Modern Greek & 1/24 & French & 0/24 & Filipino & 0/24 \\
16 & Russian & 1/24 & Ukrainian & 0/24 & Italian & 0/24 \\
17 & Ukrainian & 1/24 & Russian & 0/24 & Serbian-Croatian-Bosnian & 0/24 \\
18 & Standard Malay & 0/24 & Italian & 0/24 & Russian & 0/24 \\
19 & Serbian-Croatian-Bosnian & 0/24 & Swedish & 0/24 & Ukrainian & 0/24 \\
20 & Korean & 0/24 & Korean & 0/24 & Welsh & 0/24 \\
21 & Japanese & 0/24 & Japanese & 0/24 & Korean & 0/24 \\
22 & Welsh & 0/24 & Welsh & 0/24 & Japanese & 0/24 \\
23 & Mandarin Chinese & 0/24 & Southern Pashto & 0/24 & Mandarin Chinese & 0/24 \\
24 & Southern Pashto & 0/24 & Mandarin Chinese & 0/24 & Southern Pashto & 0/24 \\
\bottomrule
\end{tabular}}
\end{table}

\FloatBarrier

\paragraph{Residual adjustment detail}
The bias layer explains $R^2{=}0.12$ of BTS in-sample but roughly zero under LOLO, which is why its weights describe ATLAS-24 rather than a stable law.


\end{document}